%% file: main.tex
\documentclass[acmtog]{acmart}
\acmSubmissionID{0875}

\usepackage{booktabs} 

\usepackage[ruled]{algorithm2e} 

\SetAlFnt{\small}
\SetAlCapFnt{\small}
\SetAlCapNameFnt{\small}
\SetAlCapHSkip{0pt}

\copyrightyear{2023}
\acmYear{2023}
\setcopyright{acmlicensed}\acmConference[SIGGRAPH '23 Conference Proceedings]{Special Interest Group on Computer Graphics and Interactive Techniques Conference Conference Proceedings}{August 6--10, 2023}{Los Angeles, CA, USA}
\acmBooktitle{Special Interest Group on Computer Graphics and Interactive Techniques Conference Conference Proceedings (SIGGRAPH '23 Conference Proceedings), August 6--10, 2023, Los Angeles, CA, USA}
\acmPrice{15.00}
\acmDOI{10.1145/3588432.3591566}
\acmISBN{979-8-4007-0159-7/23/08}

\usepackage{graphicx}
\usepackage{amsmath}
\usepackage{mathtools}
\usepackage{booktabs}
\usepackage{comment}
\usepackage{xcolor}
\usepackage{float}
\usepackage{subcaption}

\DeclareMathOperator*{\argmin}{arg\,min}

\newcommand{\etal}{\textit{et al.}}
\newcommand{\OURS}{ClipFace}

\newcommand{\todo}[1]{{\textcolor{blue}{#1}}}

\begin{document}
\title{\OURS: Text-guided Editing of Textured 3D Morphable Models}

\author{Shivangi Aneja}
\affiliation{%
 \institution{Technical University of Munich}
 \country{Germany}}
 \email{shivangi.aneja@tum.de}

\author{Justus Thies}
\affiliation{%
 \institution{Max Planck Institute for Intelligent Systems, T{\"u}bingen}
  \country{Germany}}

 \author{Angela Dai}
\affiliation{%
 \institution{Technical University of Munich}
  \country{Germany}}

\author{Matthias Niessner}
\affiliation{%
 \institution{Technical University of Munich}
  \country{Germany}}

\input{pages/00_abstract}

\begin{teaserfigure}
  \includegraphics[width=1.0\linewidth]{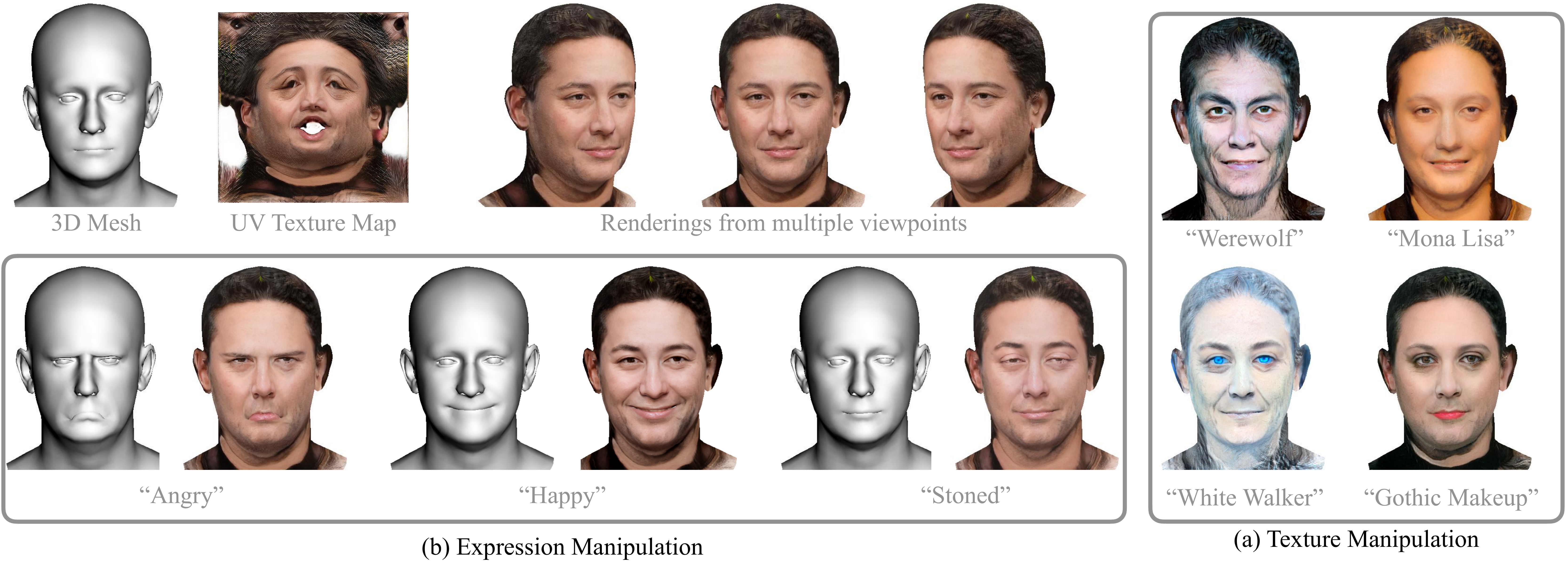}
  \caption{\OURS{} learns a self-supervised generative model for jointly synthesizing geometry and texture leveraging 3D morphable face models, that can be guided by text prompts.
    For a given 3D mesh with fixed topology, we can generate arbitrary face textures as UV maps (top).
    The textured mesh can then be manipulated with text guidance to generate diverse set of textures and geometric expressions in 3D by altering (a) only the UV texture maps for Texture Manipulation and (b) both UV maps and mesh geometry for Expression Manipulation.}
    \label{fig:teaser}
\end{teaserfigure}

\maketitle
\input{pages/01_intro}
\input{pages/02_related_work}
\input{pages/03_method}
\input{pages/04_results}
\input{pages/05_conclusion}

\clearpage
\input{pages/06_appendix}

\bibliographystyle{ACM-Reference-Format}
\bibliography{main}

\end{document}

%% file: pages/00_abstract.tex
\begin{abstract}
We propose \OURS, a novel self-supervised approach for text-guided editing of textured 3D morphable model of faces.
Specifically, we employ user-friendly language prompts to enable control of the expressions as well as appearance of 3D faces.
We leverage the geometric expressiveness of 3D morphable models, which inherently possess limited controllability and texture expressivity, and develop a self-supervised generative model to jointly synthesize expressive, textured, and articulated faces in 3D.
We enable high-quality texture generation for 3D faces by adversarial self-supervised training, guided by differentiable rendering against collections of real RGB images.
Controllable editing and manipulation are given by language prompts to adapt texture and expression of the 3D morphable model.
To this end, we propose a neural network that predicts both texture and expression latent codes of the morphable model.
Our model is trained in a self-supervised fashion by exploiting differentiable rendering and losses based on a pre-trained CLIP model.
Once trained, our model jointly predicts face textures in UV-space, along with expression parameters to capture both geometry and texture changes in facial expressions in a single forward pass. 
We further show the applicability of our method to generate temporally changing textures for a given animation sequence.
\end{abstract}

%% file: pages/01_intro.tex
\section{Introduction}

Modeling 3D content is central to many applications in our modern digital age, including asset creation for video games and films, as well as mixed reality.
In particular, modeling 3D human face avatars is a fundamental element towards digital expression.
However, current content creation processes require extensive time from highly-skilled artists in creating compelling 3D face models.

In contrast to implicit representations for human faces~\cite{Chan2021} that do not follow fixed mesh topology, 3D morphable models present a promising approach for modeling animatable avatars, with popular blendshape models used for human faces (e.g., FLAME~\cite{flame_siggraphAsia2017}) or  bodies (e.g., SMPL~\cite{SMPL_2015}).
In particular, they offer a compact, parametric representation to model an object, while maintaining a mesh representation that fits the classical graphics pipelines for editing and animation.
Additionally, the shared topology of the representation enables deformation and texture transfer capabilities.

Despite such morphable models' expressive capability in geometric modeling and potential practical applicability towards artist creation pipelines, they remain insufficient for augmenting artist workflows.
This is due to limited controllability, as they rely on PCA models, and lack of texture expressiveness, since the models have been built from very limited quantities of 3D-captured textures; both of these aspects are crucial for content creation and visual consumption.
We thus address the challenging task of creating a generative model to enable synthesis of expressive, textured, and articulated human faces in 3D.

We propose \OURS{}, to enable controllable generation and editing of 3D faces.
We leverage the geometric expressiveness of 3D morphable models, and introduce a self-supervised generative model to jointly synthesize textures and adapt expression  parameters of the morphable model.
To facilitate controllable editing and manipulation, we exploit the power of vision-language models~\cite{clip_radford21a} to enable user-friendly generation of diverse textures and expressions in 3D faces.
This allows us to specify facial expressions as well as the appearance of the human via text while maintaining a clean 3D mesh representation that can be consumed by standard graphics applications.
Such text-based editing enables intuitive control over the content creation process.

Our generative model is trained in a self-supervised fashion, leveraging the availability of large-scale face image datasets with differentiable rendering to produce a powerful texture generator that can be controlled along with the morphable model geometry by text prompts. Based on our texture generator, we learn a neural network that can edit the texture latent code as well as the expression parameters of the 3D morphable model with text prompt supervision and losses based on CLIP.
Our approach further enables generating temporally varying textures for a given driving expression sequence.

\medskip
\noindent
The main focus of our work lies in enabling text-guided editing and control of 3D morphable face models. As recent 3D face models~\cite{flame_siggraphAsia2017,bfm_17} are limited in their texture space, we propose a generative model to synthesize UV textures with a realistic appearance; this texture space is a pre-requisite for our text-guided editing. To summarize, we present the following contributions:

%
\begin{itemize}
    \item We propose a novel approach to controllable editing of textured, parametric 3D morphable models through user-friendly text prompts, by exploiting CLIP-based supervision to jointly synthesize texture and expressions of a 3D face model.
    \item The controllable 3D face model is supported by our texture generator, trained in a self-supervised fashion on 2D images only.
    \item Our approach additionally enables generating temporally varying textures of an animated 3D face model from a driving video sequence.
\end{itemize}

%% file: pages/02_related_work.tex
\section{Related Work}

\textit{\textbf{Texture Generation:}}
There is a large corpus of research works in the field of generative models for UV textures~\cite{gecer2020tbgan, Gecer_2019_CVPR, gecer2021fast, Gecer_2021_CVPR, lattas2020avatarme, lattas2021avatarme++, Luo_2021_CVPR, Lee2020StyleUVDA, Li_2020_CVPR,Wang_2022_CVPR}.
These methods achieve impressive results; however, the majority is fully supervised in nature, requiring ground truth textures, which in turn necessitate collection in a controlled capture setting.
Learning self-supervised texture generation is much more challenging, and only a handful of methods exist.
For instance, Marriott \etal~\shortcite{Marriott2021A3G} were among the first to leverage Progressive GANs~\cite{karras2018progressive} and 3D Morphable Models~\cite{3dmm} to generate textures for facial recognition; however, the textures generated remain relatively low resolution and are not suitable to perform language-driven texture and expression manipulations.

Textures generated by Slossberg \etal~\shortcite{slossberg2021unsupervised} made a significant improvement in quality by using pretrained StyleGAN~\cite{Karras2019stylegan2} and StyleRig~\cite{tewari2020stylerig}.
The closest inspiration to our texture generator is StyleUV~\cite{Lee2020StyleUVDA} which also operates on a mesh. 
Both methods achieve stunning results, but  do not take the head and ears into account, which limits their practical applicability to directly use them as assets in games and movies.
In our work, we propose a generative model to synthesize UV textures for the full-head topology to enable text-guided editing and control of 3D morphable face models.
%

\textit{\textbf{Semantic Manipulation of Facial Attributes:}}
Facial manipulation has also seen significant studies following the success of StyleGAN2~\cite{Karras2019stylegan2} image generation.
In particular, its disentangled latent space facilitates texture editing as well as enables a level of control over pose and expressions of generated images.
Several methods~\cite{KowalskiECCV2020, deng2020disentangled, GIF2020, tewari2020stylerig, tewari2020pie, Liu20223DFMGT, styleflow} have made significant progress to induce controllability to images by embedding 3D priors via conditioning StyleGAN on known facial attributes extracted from synthetic face renderings.
However, these methods operate on 2D images, and although they achieve high-quality results on a per-frame basis, consistent and coherent rendering from varying poses and expressions remains challenging.
Motivated by such impressive 2D generators, we propose to lift them to 3D and directly operate in UV space of a 3D mesh, producing temporally-consistent results when rendering animation sequences. 
%

\textit{\textbf{Text-Guided Image Manipulation:}}
Recent progress in 2D language models has opened up significant opportunities for text-guided image manipulation~\cite{avrahami2022blended,ruiz2022dreambooth, clip2stylegan, bau2021paintbyword, vqgan_clip, Dayma_dalle_Mini_2021, dalle, dalle2}. 
For instance, contrastive language-image pre-training (CLIP)~\cite{clip_radford21a} model has been used for text-guided editing for a variety of 2D/3D applications~\cite{Patashnik_2021_ICCV, gal2021stylegannada, Michel_2022_CVPR, Kocasari_2022_WACV, latent_3d, petrovich22temos, hong2022avatarclip, youwang2022clipactor, Wei_2022_CVPR, Wang_2022_CVPR_clip_nerf, khalid2022clipmesh}.
StyleClip~\cite{Patashnik_2021_ICCV} presented seminal advances in stylizing human face images by leveraging the expressive power of CLIP in combination with the generative power of StyleGAN to produce unique manipulations for faces.
This was followed by StyleGAN-Nada~\cite{gal2021stylegannada}, which enables adapting image generation to a remarkable diversity of styles from various domains, without requiring image examples from those domains.
However, these manipulations are designed for the image space, and are not 3D consistent. 
%

\begin{figure*}[t!]
    \begin{center}
    \includegraphics[width=1.0\linewidth]{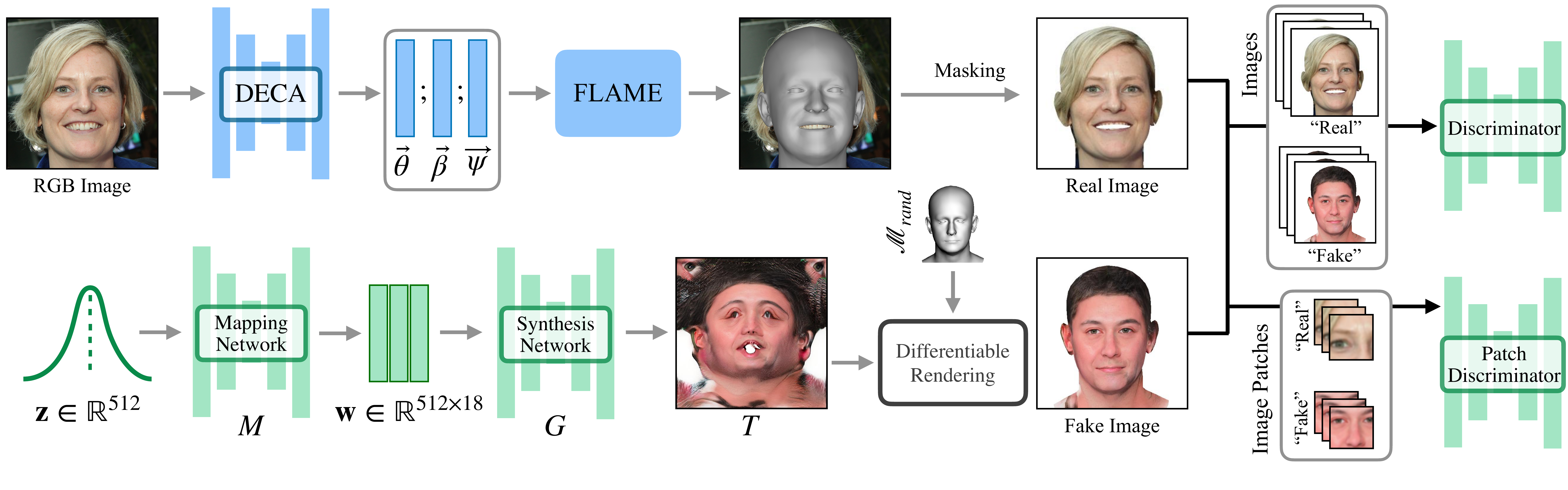}
    \end{center}
    \vspace{-0.6cm}
      \caption{
      \textbf{Texture Generation:}
      We learn self-supervised texture generation from collections of 2D images. An RGB image is encoded by pretrained encoder DECA~\cite{deca} to extract shape $\boldsymbol{\vec{{\beta}}}$, pose $\boldsymbol{\vec{{\theta}}}$ and expression $\boldsymbol{\vec{{\psi}}}$ coefficents in FLAME's latent space, which are decoded by FLAME~\cite{flame_siggraphAsia2017} to deformed mesh vertices. 
      The background and mouth interior are then masked out, generating the `Real Image' for our adversarial formulation. 
      In parallel, latent code $z \in \mathbb{R}^{512}$ is sampled from a gaussian distribution $\mathcal{N}(0,I)$ and input to mapping network $M$ to generate intermediate latent $\mathbf{w} \in \mathbb{R}^{512 \times 18}$, which is used by synthesis network $G$ to generate the UV texture image. The predicted texture is differentiably rendered on a randomly deformed FLAME mesh  to generate the `Fake Image.' Two discriminators interpret the generated and masked real image, at full resolution and at patch size $64 \times 64$. Frozen models are denoted in blue, and learnable networks in green.
      }
    \label{fig:texture_gen}
\end{figure*}

\textit{\textbf{Text-Guided 3D Manipulation:}}
Following the success of text-guided image manipulation, recent works have adopted powerful vision-language models to enable text guidance for 3D object manipulation.
Text2Mesh~\cite{Michel_2022_CVPR} was one of the first pioneering methods to leverage a pre-trained 2D CLIP model as guidance to generate language-conditioned 3D mesh textures and geometric offsets.
Here, edits are realized as part of a test-time optimization that aims to solve for the texture and mesh offsets in a neural field representation, such that their re-renderings minimizes a 2D CLIP loss from different viewpoints.
Similar to Text2Mesh, CLIP-Mesh~\cite{khalid2022clipmesh} produces textured meshes by jointly estimating texture and deformation of a template mesh, based on text inputs using CLIP.
Recently, Canfes \etal~\shortcite{latent_3d} adapt TB-GAN~\cite{gecer2020tbgan}, an expression-conditioned generative model to produce UV-texture maps, with a CLIP loss to produce facial expressions in 3D, although the quality of textures and expressions is limited, due to reliance on 3D scan data for TB-GAN training.
Our method is inspired by these lines of research; however, our focus lies on leveraging the parametric representations of 3D morphable face models with our StyleGAN-based texture generator, which can enable content creation for direct use in many applications such as games or movies.

%% file: pages/03_method.tex
\section{Method}

ClipFace targets text-guided synthesis of textured 3D face models.
It consists of two fundamental components:
(i) an expressive generative texture space for facial appearances (Sec~\ref{subsec:tex_gen}),
and (ii) a text-guided prediction of the latent codes for the texture generator and the expression parameters of the underlying statistical morphable model (Sec~\ref{subsec:tex_exp_manipulation}).
In the following, we will detail these contributions and further demonstrate how they enable producing temporally changing textures for a given animation sequence Sec.~\ref{subsec:video_animation}.
%

\begin{figure*}[tp]
    \begin{center}
    \includegraphics[width=1.0\linewidth]{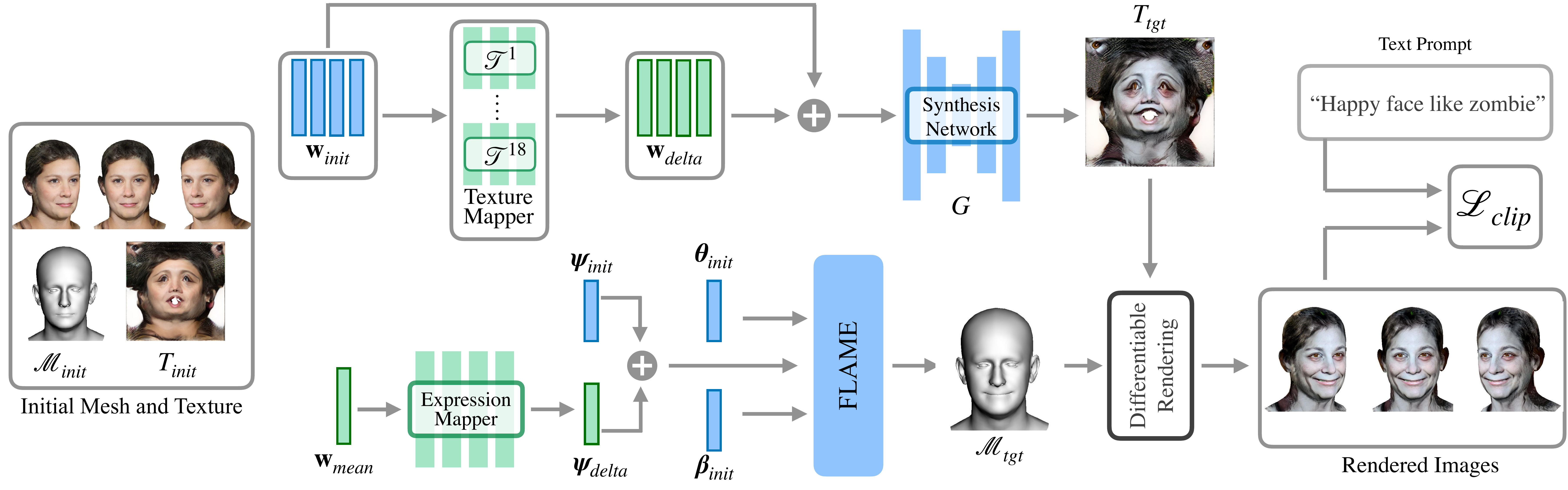}
    \end{center}
    \vspace{-0.5cm}
      \caption{
      \textbf{Text-guided Synthesis of Textured 3D Face Models:}
      From a given textured mesh with texture code $\mathbf{w}_{\textrm{init}}$, we synthesize various styles by adapting both texture and expression to the target text prompt.
      $\mathbf{w}_{\textrm{init}}$ is input to the texture mappers $\mathcal{T} = [\mathcal{T}^1, ..., \mathcal{T}^{18}]$ to obtain texture offsets $\mathbf{w}_{\textrm{delta}} \in \mathbb{R}^{512 \times 18}$ for 18 different levels of  $\mathbf{w}_{\textrm{init}}$. The expression mapper $\mathcal{E}$ takes mean latent code $\mathbf{w}_{\textrm{mean}} = \| \mathbf{w}_{\textrm{init}} + \mathbf{w}_{\textrm{delta}} \|_2$ as input, and predicts expression offset $\boldsymbol{\psi}_{\textrm{delta}}$ to obtain deformed mesh geometry $\mathcal{M}_{\textrm{tgt}}$. The generated UV map $T_{\textrm{tgt}}$ and deformed mesh $\mathcal{M}_{\textrm{tgt}}$ are differentiably rendered to generate styles that fit the text prompt.
      }
    \label{fig:clip_learning}
\end{figure*}

\subsection{Generative Synthesis of Face Appearance}
\label{subsec:tex_gen}
 
Since there does not exist any large-scale datasets for UV textures, we propose a self-supervised method to learn the appearance manifold of human faces, as depicted in Fig.~\ref{fig:texture_gen}.
Rather than learning from ground truth UV textures, we instead leverage large-scale RGB image datasets of faces, which we use in an adversarial formulation through differentiable rendering.
{For our experiments, we use the FFHQ dataset~\cite{Karras2018stylegan} which consists of 70,000 diverse, high-quality images.
As we focus on the textures of the human head, we remove images that contain headwear (caps, scarfs, etc.) and eyewear (sunglasses and spectacles) using face parsing~\cite{face_parsing_pytorch}, resulting in a subset of 45,000 images.
Based on this data, we train a StyleGAN-ADA~\cite{Karras2020ada} generator to produce UV textures that when rendered on top of the FLAME mesh~\cite{flame_siggraphAsia2017} results in realistic imagery.
More specifically, we use the FLAME model as our shape prior to produce different geometric shapes and facial expressions.
It can be defined as:
\begin{equation}
    \mathcal{F(\boldsymbol{\vec{\beta}}, \boldsymbol{\vec{\theta}}, \boldsymbol{\vec{\psi}}): \mathbb{R}^{|\boldsymbol{\vec{\beta}}| \times |\boldsymbol{\vec{\theta}}| \times |\boldsymbol{\vec{\psi}}|}} \rightarrow \mathbb{R}^{3N},
\end{equation}
where $\boldsymbol{\vec{\beta}} \in \mathbb{R}^{100}$ are the shape parameters, $\boldsymbol{\vec{\theta}} \in \mathbb{R}^{6}$ refers to the jaw and head pose, and $\boldsymbol{\vec{\psi}} \in \mathbb{R}^{50}$ are the expression coefficients.
To recover the distribution of face shapes and expressions from the training dataset, we employ DECA~\cite{deca}, a pretrained encoder that takes an RGB image as input and outputs the corresponding FLAME parameters $\boldsymbol{\vec{\beta}}, \boldsymbol{\vec{\theta}}, \boldsymbol{\vec{\psi}}$, including orthographic camera parameters $\mathbf{c}$.
We use the recovered parameters to remove the backgrounds from the original images, and only keep the image region that is covered by the corresponding face model.
Using this distribution of face geometries and camera parameters $\mathcal{D} \sim [\boldsymbol{\vec{\beta}}, \boldsymbol{\vec{\theta}}, \boldsymbol{\vec{\psi}}, \mathbf{c}]$, along with the masked real samples, we train the StyleGAN network using differentiable rendering~\cite{Laine2020diffrast}.
We sample a latent code $\boldsymbol{z} \in \mathbb{R}^{512}$ from Gaussian distribution $\mathcal{N(\boldsymbol{0},\bold{I})}$ to generate the intermediate latent code $\mathbf{w} \in \mathbb{R}^{512 \times 18}$ using the a mapping network $M$: $\mathbf{w} = M(\mathbf{z})$.
This latent code $\mathbf{w}$ is passed to the synthesis network $G$ to generate UV texture map $T \in \mathbb{R}^{512 \times 512 \times 3}$: $T = G(\mathbf{w})$.
This predicted texture $T$ is then rendered on a randomly sampled deformed mesh from our discrete distribution of face geometries $\mathcal{D}$.
We use an image resolution of $512 \times 512$.
Both the generated image and masked real image are then passed to the discriminator during training.
To further improve the texture quality, we use a patch discriminator alongside a full-image discriminator, which encourages high-fidelity details in local regions of the rendered images.
We apply image augmentations (e.g., color jitter, image flipping, hue/saturation changes) to both full-image and image patches before feeding them to the discriminator.
The patch size is set to $64 \times 64$ for all of our experiments.
Note that the patch discriminator is critical to producing high-frequency texture details; see Section ~\ref{sec:results}.

\begin{figure*}[tp]
    \begin{center}
    \includegraphics[width=1.0\linewidth]{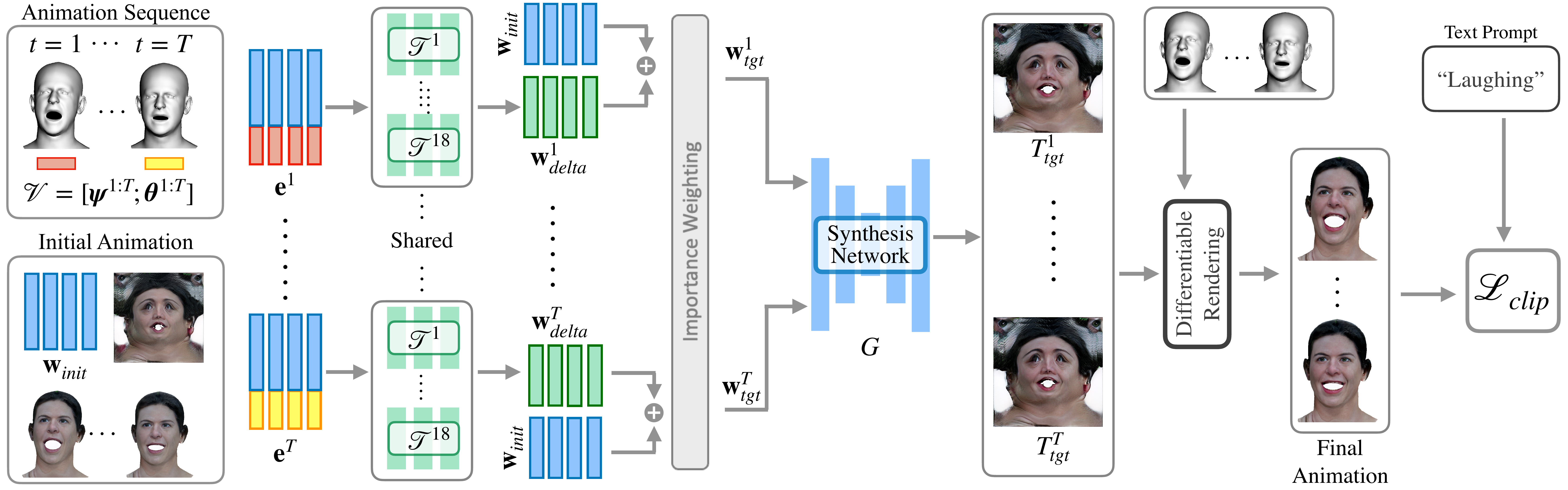}
    \end{center}
    \vspace{-0.3cm}
      \caption{
      \textbf{Texture Manipulation for Animation Sequences:}
      Given a video sequence $\mathcal{V} = [\boldsymbol{\psi}^{1:T}; \boldsymbol{\theta}^{1:T}]$ of T frames, an initial texture code $\mathbf{w}_{\textrm{init}}$, and a text prompt, we synthesize a 3D textured animation to match the text. We concatenate $\mathcal{V}$ to $\mathbf{w}_{\textrm{\textrm{init}}}$ across different timestamps to obtain $\mathbf{e}^t$, which is input to the time-shared texture mappers $\mathcal{T}$ to obtain time-dependent textures offsets $\mathbf{w}_{\textrm{delta}}^{1:T}$ for all frames. The new texture codes $\mathbf{w}_{\textrm{tgt}}^{1:T}$ generated using importance weighting, are then passed to our texture generator $G$ to obtain time-dependent UV textures $T_{\textrm{tgt}}^{1:T}$, which are then differentiably rendered to generate the final animation, guided by the CLIP loss across all frames.
      }
    \label{fig:clip_learning_video}
\end{figure*}

\subsection{Text-guided Synthesis of Textured 3D Models}\label{subsec:tex_exp_manipulation}

For a given textured mesh with texture code $\mathbf{w}_{\textrm{init}} = \{\mathbf{w}_{\textrm{init}}^1,  ... \mathbf{w}_{\textrm{init}}^{18} \} \in \mathbb{R}^{512 \times 18}$ in neutral pose $\boldsymbol{\theta}_{\textrm{init}}$ and neutral expression $\boldsymbol{\psi}_{\textrm{init}}$, our goal is to learn optimal offsets $\mathbf{w}_\textrm{{delta}}, \boldsymbol{\psi}_{\textrm{delta}}$ for texture and expression respectively defined through text prompts.
As a source of supervision, we use a pretrained CLIP~\cite{clip_radford21a} model due to its high expressiveness, and formulate the offsets as:
\begin{equation}
    \mathbf{w}_{\textrm{delta}}^{*}, \boldsymbol{\psi}_{\textrm{delta}}^{*} = \argmin_{\mathbf{w}_{\textrm{delta}}, \boldsymbol{\psi}_{\textrm{delta}}} \mathcal{L}_{\textrm{total}},
\end{equation}
where $\mathcal{L}_{\textrm{total}}$ formulates CLIP guidance and expression regularization, as defined in Eq.~\ref{eq:totalloss}.
In order to optimize this loss, we learn a texture mapper $\mathcal{T} = [\mathcal{T}^1, ..., \mathcal{T}^{18}]$  and an expression mapper $\mathcal{E}$.
The texture mapper predicts the latent variable offsets across the different levels $\{1, 2, ... 18 \} $ of the StyleGAN generator:
\begin{equation}
    \mathbf{w}_{\textrm{delta}}= 
    \begin{cases}
        \mathcal{T}^{1}(\mathbf{w}_\textrm{{init}}^{1})\\
        \mathcal{T}^{2}(\mathbf{w}_{\textrm{init}}^{2})\\
        \vdots\\
        \mathcal{T}^{18}(\mathbf{w}_{\textrm{init}}^{18}).\\
    \end{cases}
\end{equation}
The expression mapper $\mathcal{E}$ learns the expression offsets and takes as input $\mathbf{w}_{\textrm{mean}} = \| \mathbf{w}_{\textrm{init}} + \mathbf{w}_{\textrm{delta}} \|_2$, where $\mathbf{w}_{\textrm{mean}} \in \mathbb{R}^{512}$ is the mean of 18-different levels of the latent space, and outputs the expression offsets $\boldsymbol{\psi}_{\textrm{delta}}$:
\begin{equation}
    \begin{aligned}
        \boldsymbol{\psi}_{\textrm{delta}} = \mathcal{E}(\mathbf{w}_{\textrm{mean}}).
    \end{aligned}
\end{equation}
We notice that it is critical to design separate texture and expression mappers to maintain disentangled texture and expression spaces. Conditioning the expression mapper on texture codes correlates them meaningfully for realistic expression generation, significantly improving generation quality. We show results for different input conditions in supplemental. We use a 4-layer MLP architecture with LeakyReLU activations for the mappers. The method is shown in Fig.~\ref{fig:clip_learning}.
Naively using a CLIP loss as in StyleClip~\cite{Patashnik_2021_ICCV} to train the mappers tends to result in unwanted identity and/or illumination changes in texture.
Thus, we draw inspiration from ~\cite{gal2021stylegannada}, and leverage the CLIP-space direction between the initial style and the to-be-performed manipulation in order to perform consistent and identity-preserving manipulation.
We compute the `text-delta' direction $\Delta \mathbf{t}$ in CLIP-space between the initial text prompt $\mathbf{t}_{\textrm{init}}$ and the target text prompt $\mathbf{t}_{\textrm{tgt}}$, indicating which attributes from the initial style should be changed:
\begin{equation}
  \Delta \mathbf{t} = E_{T}(\mathbf{t}_{\textrm{tgt}}) - E_{T}(\mathbf{t}_{\textrm{init}}),
\end{equation}
where  $E_{T}$ refers to the CLIP text encoder. We use the same initial text $\mathbf{t}_{\textrm{init}}$ = `A photo of a face' for all our experiments and alter the target text prompt $\mathbf{t}_{\textrm{tgt}}$, depending on the desired style change. For example, to generate a Mona Lisa style texture, we use the text prompt $\mathbf{t}_{\textrm{tgt}}$ = `A photo of a face that looks like Mona Lisa'.
Guided via the CLIP-space image direction between the initial rendered image $\mathbf{i}_{\textrm{init}}$ and the target rendered image $\mathbf{i}_{\textrm{tgt}}$ generated using our texture generator, we train the mapping networks to predict style specified by the target text prompt $\mathbf{t}_{\textrm{tgt}}$.
\begin{equation}
    \Delta \mathbf{i} = E_{I}(\mathbf{i}_{\textrm{tgt}}) - E_{I}(\mathbf{i}_{\textrm{init}}),
\end{equation}
where $\mathbf{i}_{\textrm{init}}$ is the image rendered with the initial texture and initial flame parameters $\left(\mathbf{w}_{\textrm{init}}, \boldsymbol{\beta}, \boldsymbol{\theta}, \boldsymbol{\psi}_{\textrm{init}}\right)$, and $\mathbf{i}_{\textrm{tgt}}$ is the image with the target texture and target flame parameters $\left(\mathbf{w}_{\textrm{tgt}}, \boldsymbol{\beta}, \boldsymbol{\theta}, \boldsymbol{\psi}_{\textrm{tgt}}\right)$, and $E_{I}$ refers to the CLIP image encoder.

Note that we do not alter the pose $\boldsymbol{\theta}$ and shape code $\boldsymbol{\beta}$ of the FLAME model.
The CLIP loss $\mathcal{L}_{\textrm{clip}}$ is then computed as:
\begin{equation}
    \mathcal{L}_{\textrm{clip}} = 1- \frac{\Delta \mathbf{i} .  \Delta \mathbf{t}}{|\Delta \mathbf{i}| \dot  |\Delta \mathbf{t}|}.
    \label{eq:clip_loss}
\end{equation}
In order to prevent the mesh from taking unrealistic expressions, we further regularize the expressions using the Mahalanobis prior as:
\begin{equation}
\label{eq:tex_reg}
    \mathcal{L}_{\textrm{reg}} = \boldsymbol{\psi}^T \Sigma_{\psi}^{-1}\boldsymbol{\psi},
\end{equation}
where $\Sigma_{\psi}^{-1}$ is the diagonal expression covariance matrix of FLAME model. As we show in our results, this regularization is critical to prevent the 3D morphable model from taking an unrealistic shape.

The full training loss can then be written as:
\begin{equation}\label{eq:totalloss}
    \mathcal{L}_{\textrm{total}} = \mathcal{L}_{\textrm{clip}} + \lambda_{\textrm{reg}} \mathcal{L}_{\textrm{reg}}.
\end{equation}
Note that we can also only alter the texture without changing expressions by keeping the expression mapper frozen and not fine-tuning it. We pre-train the mapper networks to predict zero-offsets (details in supplemental).
%


\subsection{Texture Manipulation for Video Sequences}\label{subsec:video_animation}

Given an expression video sequence, we propose a novel technique to manipulate the textures for every frame of the video guided by a CLIP loss (see Fig.~\ref{fig:clip_learning_video}).
That is, for a given animation sequence $\mathcal{V} = [\boldsymbol{\theta}^{1:T}; \boldsymbol{\psi}^{1:T} ]$ of $T$ frames, with expression codes $\boldsymbol{\psi}^{1:T} = [\boldsymbol{\psi}^1, \boldsymbol{\psi}^2, ...\boldsymbol{\psi}^T]$, pose codes $\boldsymbol{\theta}^{1:T} = [\boldsymbol{\theta}^1, \boldsymbol{\theta}^2, ...\boldsymbol{\theta}^T]$, and a given texture code $\mathbf{w}_{\textrm{\textrm{init}}}$, we use a multi-layer perceptron as our texture mapper $\mathcal{T} = [\mathcal{T}^1, ..., \mathcal{T}^{18}]$ to generate time-dependent texture offsets $\mathbf{w}_{\textrm{delta}}^{1:T}$ for different levels of the texture latent space.
This mapper receives as input $\mathbf{e}^{1:T}$, the concatenation of the initial texture code $\mathbf{w}_{\textrm{\textrm{init}}}$ with the time-dependent expression and pose code $[\boldsymbol{\psi}^t; \boldsymbol{\theta}^t]$.
Mathematically, we have:
\begin{gather} 
    \mathbf{e}^{1:T} = [\mathbf{e}^{1}, .... \mathbf{e}^{T}]  \\ 
    \mathbf{e}^{t} = [\mathbf{w}_{\textrm{init}}; \boldsymbol{\psi}^t;  \boldsymbol{\theta}^t],
\end{gather}
where $ \boldsymbol{\psi}^t$ and $\boldsymbol{\theta}^t$ refer to the expression and pose code at timestamp $t$ extracted from sequence $\mathcal{V}$. Next, we pass $\mathbf{e}^{1:T}$ to the time-shared texture mapper $\mathcal{T}$ to obtain texture offsets $\mathbf{w}^{1:T}_{\textrm{delta}}$. To ensure a coherent animation and smooth transition across frames, we weight the predicted offsets $\mathbf{w}_{\textrm{delta}}^{1:T}$ using importance weights $\mathcal{I} = [i_1, ....i_T]$ extracted from video sequence $\mathcal{V}$, before adding them to $\mathbf{w}_{\textrm{\textrm{init}}}$:
\begin{equation}
    \mathbf{w}^{t}_{\textrm{\textrm{tgt}}} = \mathbf{w}_{\textrm{\textrm{init}}} +  i_t \cdot \mathbf{w}^{t}_{\textrm{\textrm{delta}}}.
\end{equation}
We compute importance weights by measuring the deviation between the neutral shape $[\boldsymbol{\theta}_{\textrm{neutral}}; \boldsymbol{\psi}_{\textrm{neutral}}]$ and per-frame face shape $[\boldsymbol{\theta}^{t}; \boldsymbol{\psi}^{t}]$, with by min-max normalization:
\begin{equation}
    i_t = \frac{\delta^t - \text{min}(\boldsymbol{\delta}^{1:T})}{\text{max}(\boldsymbol{\delta}^{1:T}) - \text{min}(\boldsymbol{\delta}^{1:T})}, 
\end{equation}
with ${\delta}^{t} = \big|\big| [\boldsymbol{\theta}_{\textrm{neutral}}; \boldsymbol{\psi}_{\textrm{neutral}}] - [\boldsymbol{\theta}^{t}; \boldsymbol{\psi}^{t}] \big|\big|_2$.
The importance weighting ensures that key frames with strong expressions are emphasized. The predicted target latent codes $\mathbf{w}_{\textrm{tgt}}^{1:T}$ are then used to generate the UV maps $T_{\textrm{tgt}}^{1:T}$, which are differentiably rendered onto the given animation sequence $\mathcal{V}$.

To train texture mapper $\mathcal{T}$, we minimize clip loss (Eq.~\ref{eq:clip_loss}) for the given text prompt $\mathbf{t}_{\textrm{tgt}}$ and the rendered frames $\mathbf{i}_{\textrm{tgt}}^{t}$  aggregated over T timesteps for all the frames from the video.

\begin{equation}
\mathbf{w}_{\textrm{delta}}^{1:T} = \argmin_{\mathbf{w}_{\textrm{delta}}^{1:T}}  \sum_{t=1}^{T} \mathcal{L}_{\textrm{clip}} (\mathbf{t}_{\textrm{tgt}}, \mathbf{i}_{\textrm{tgt}}^{t}).
\end{equation}

%% file: pages/04_results.tex
\section{Results}
\label{sec:results}
We evaluate \OURS{} on the tasks of texture generation, text-guided synthesis of textured 3D face models, and text-guided manipulation of animation sequences. 
For texture generation, we evaluate on standard GAN metrics FID and KID. We evaluate both of these metrics with respect to masked FFHQ images~\cite{Karras2018stylegan} (with background and mouth interior masked out) as ground truth, and generated textures rendered at $512 \times 512$ resolution for $\approx 45$K rendered textures and ground truth images. {For text-guided manipulation, we evaluate perceptual quality using FID \& KID and similarity to text prompt using CLIP score, which is evaluated as the cosine similarity to the text prompt using pre-trained CLIP models.} We use two different CLIP variants, `ViT-B/16' and `ViT-L/14', each on $224 \times 224$ pixels as input. We report average scores for these pre-trained variants.

\textbf{Implementation Details: } For our texture generator, we produce  $512 \times 512$ texture maps. We use an Adam optimizer with a learning rate of 2e-3, batch size 8, gradient penalty 10, and path length regularization 2 for all our experiments. We use a learning rate of 0.005 and 0.0001 for the expression and texture mappers, also using Adam. For differentiable rendering, we use NvDiffrast~\cite{Laine2020diffrast}. For the patch discriminator, we use a patch size of $64 \times 64$. We train for 300,000 iterations until convergence. For the text-guided manipulation experiments, we use the same model architecture for expression and texture mappers, a 4-layer MLP with LeakyReLU activations. For CLIP supervision, we use the pretrained `ViT-B/32' variant. For text manipulation tasks, we train for 5,000 iterations. 

\begin{figure}[tp]
    \begin{center}
    \includegraphics[width=\linewidth,trim={0 0 19cm 0},clip]{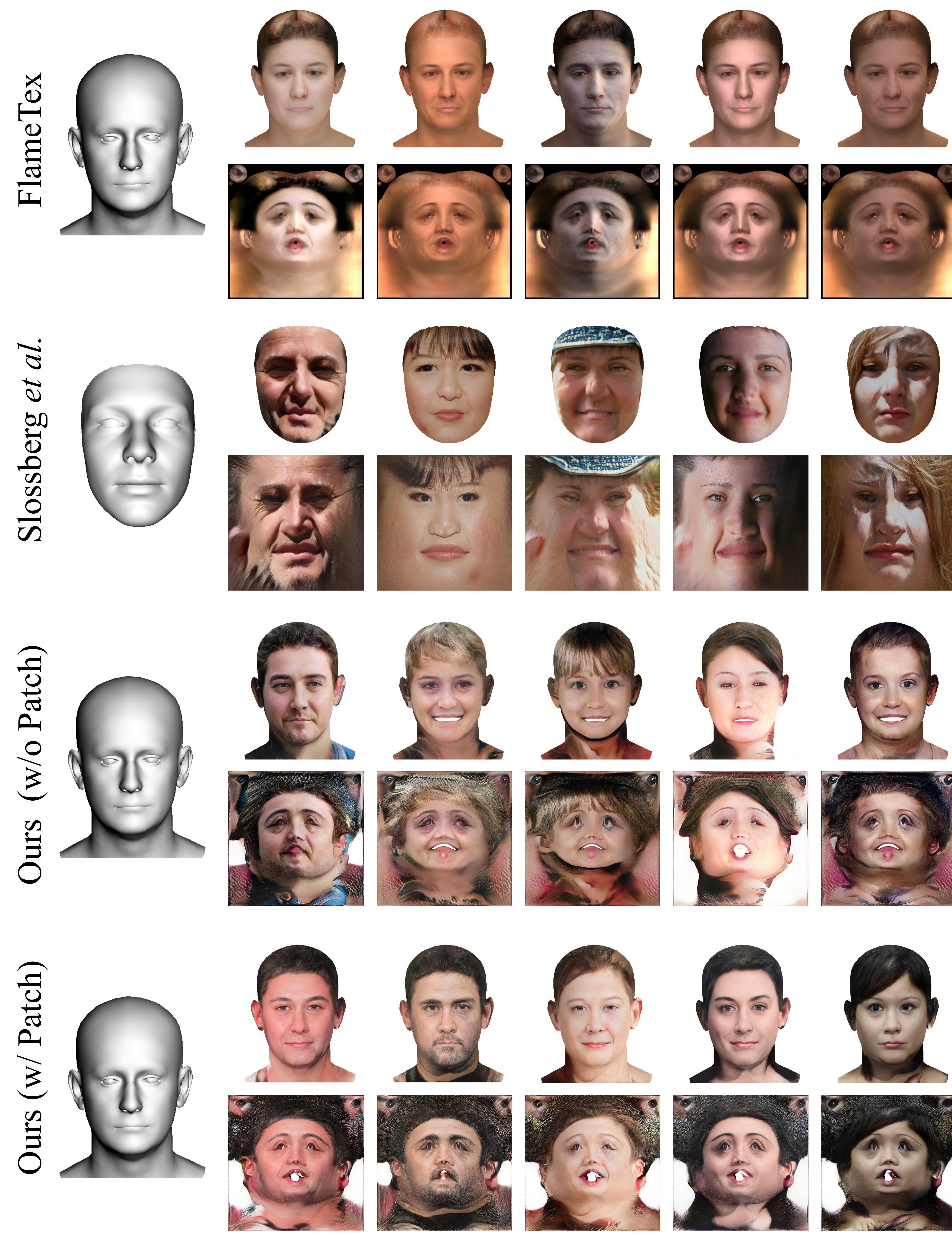}
    \end{center}
      \caption{Comparison against unsupervised texturing methods. The left column shows the mesh geometry, followed by the textured mesh and corresponding UV map. FlameTex~\cite{flame_tex} generates textures for the full head region, however, the texture quality is fairly limited. Slossberg et. al.~\shortcite{slossberg2021unsupervised} generates plausible textures, but does not take the head and ears into account,  limiting its practical applicability. Our approach is able to synthesize diverse texture styles ranging across different skin colors, ethnicities, and demographics. Note that the patch discriminator helps not only to improve texture quality but also in correctly aligning the UV texture with mesh geometry, especially around the mouth region.}
    \label{fig:stylegan_textures}
\end{figure}

\paragraph{\textbf{Texture Generation}}
We evaluate the quality of our generated textures and compare with existing unsupervised texture generation methods in Tab.~\ref{tab:stylegan_results_metrics} and Fig.~\ref{fig:stylegan_textures}.
\OURS{} outperforms other baselines in perceptual quality. Although Slossberg \etal~\shortcite{slossberg2021unsupervised} can obtain good textures for the interior face region, it does not synthesize head and ears.

\begin{table}[bp]
    \begin{center}
    \begin{tabular}{c|r|r}
        \toprule
        Method & FID $\downarrow$ & KID $\downarrow$  \\
        \toprule
        FlameTex~\cite{flame_tex} & 76.627  & 0.063 \\
        Slossberg \etal~\shortcite{slossberg2021unsupervised} & 32.794  & 0.021 \\
        Ours (w/o Patch) & 16.640 & 0.013 \\
        Ours (w/ Patch) & \textbf{9.559} & \textbf{0.006} \\
        \hline
    \end{tabular}
    \end{center}
    \caption{Quantitative evaluation of texture quality. Our approach significantly outperforms baselines in both FID and KID scores.}
    \label{tab:stylegan_results_metrics}
\end{table}

\begin{table}[bp]
    \begin{center}
    \begin{tabular}{c|l|l|l}
        \toprule
        Method & {FID} $\downarrow$ & KID $\downarrow$ & CLIP Score $\uparrow$  \\
        \toprule
        Latent3d~\cite{latent_3d} & {205.27} & 0.260 & 0.227 \textcolor{gray}{$\pm 0.041$} \\
        FlameTex~\cite{flame_tex} & {88.95} & 0.053 & 0.235 \textcolor{gray}{$\pm 0.053$}   \\
        ClipMatrix~\cite{clip_matrix} & {198.34} & 0.180  & 0.243 \textcolor{gray}{$\pm 0.049$} \\
        Text2Mesh~\cite{Michel_2022_CVPR} & {219.59} & 0.185 & \textbf{0.264} \textcolor{gray}{$\pm 0.044$} \\
        Ours & {\textbf{80.34}} & \textbf{0.032} & 0.251 \textcolor{gray}{$\pm 0.059$}  \\
        \hline
    \end{tabular}
    \end{center}
    \caption{Evaluation of text manipulation. \OURS{} effectively matches text prompts while maintaining high perceptual fidelity.}
    \label{tab:expression_results_metrics}
\end{table}

\begin{figure*}[tp]
    \begin{center}
    \includegraphics[width=.95\linewidth]{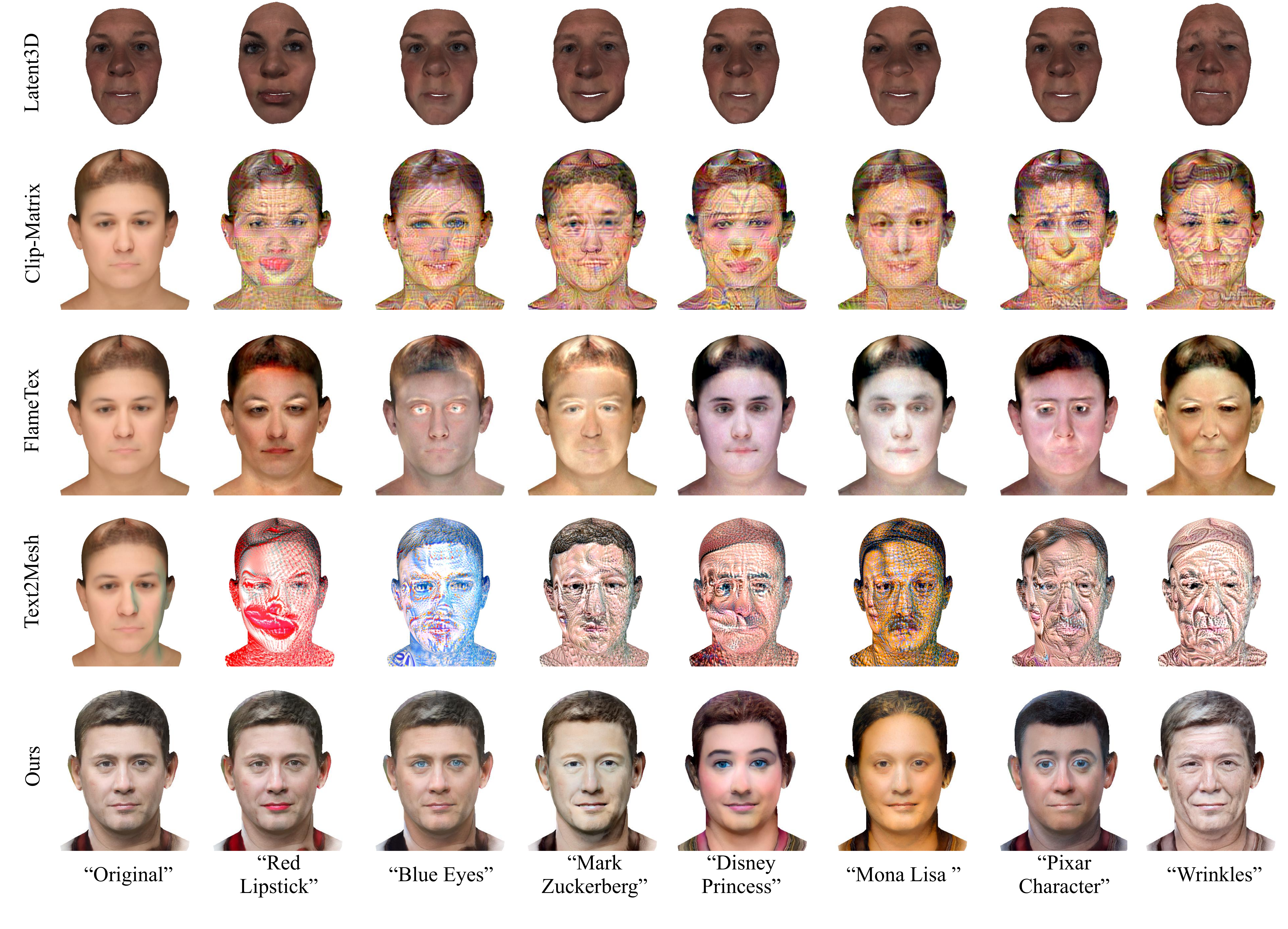}
    \end{center}
    \vspace{-0.5cm}
      \caption{Qualitative comparison on texture manipulation: We compare our method against several 3D texturing methods: Latent3D~\cite{latent_3d}, Clip-Matrix~\cite{clip_matrix}, FlameTex~\cite{flame_tex}, Text2Mesh~\cite{Michel_2022_CVPR}. Our method obtains consistently superior textures in comparison to the baselines, even capable deftly adapting identity when guided by text prompt.}
    \label{fig:clip_baseline_comp}
\end{figure*}

\paragraph{\textbf{Texture \& Expression Manipulation}}
We compare with CLIP-based texturing techniques for texture manipulation in Fig.~\ref{fig:clip_baseline_comp} and Tab.~\ref{tab:expression_results_metrics}. 
Note that for comparisons with Text2Mesh~\cite{Michel_2022_CVPR}, we follow the authors' suggestion to first perform remeshing to increase vertices from 5023 to 60,000 before optimization.
Our approach generates consistently high-quality textures for various prompts, in comparison to baselines.
In particular, our texture generator enables effective editing even in small face regions (e.g., lips and eyes).
While Text2Mesh yields a high CLIP score, it produces semantically implausible results, as the specified text prompts highly match rendered colors irrespective of the global face context (i.e., which region should be edited).
In contrast, our method generates perceptually high-quality face texture, evident in the perceptual KID metric.

\begin{figure*}
    \begin{center}
    \includegraphics[width=.9\linewidth]{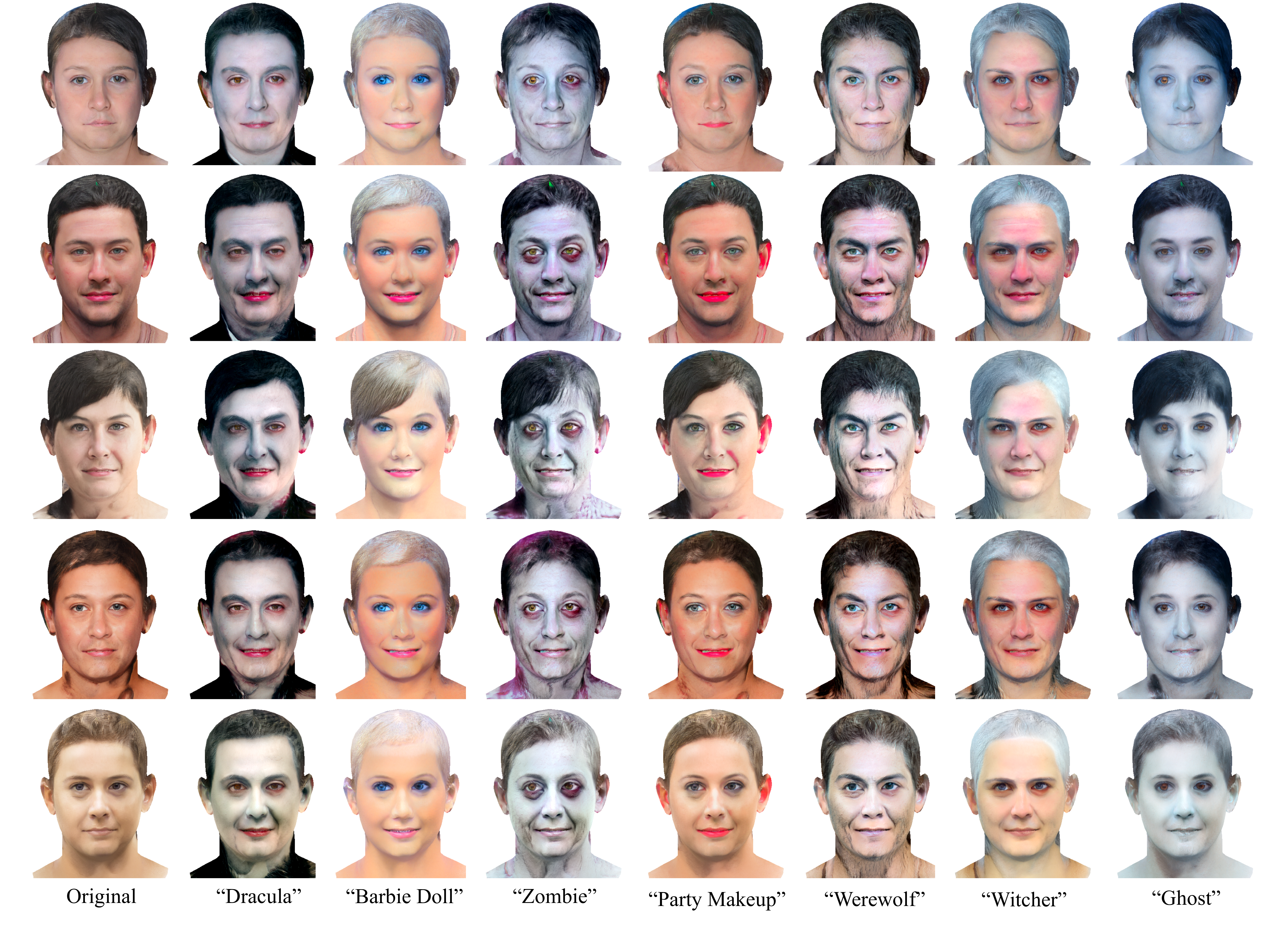}
    \end{center}
    \vspace{-0.6cm}
      \caption{
      Texture manipulations. `Original' (first column) shows the  initial textured mesh sampled from our \OURS{} generator without any text prompt, followed by \OURS{}-generated textures for various text prompts.}
    \label{fig:texturing_ours}
\end{figure*}

We show additional \OURS{} texturing results on a wide variety of prompts, including on fictional characters, in Fig.~\ref{fig:texturing_ours}, demonstrating our expressive power.

\begin{figure*}
    \begin{center}
    \includegraphics[width=1.0\linewidth]{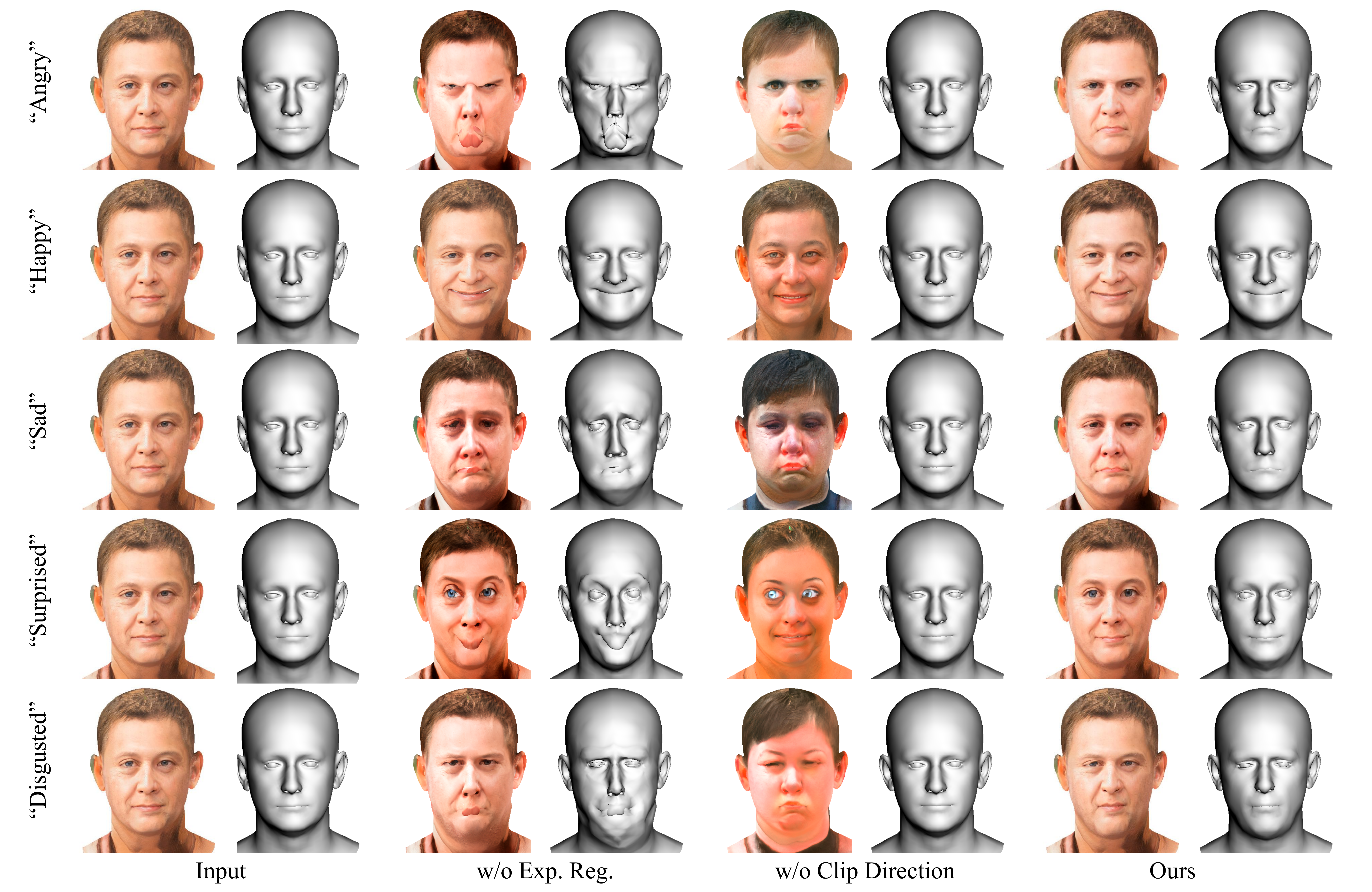}
    \end{center}
      \caption{
      \OURS{} generates a large variety of expressions, faithfully deforming the mesh geometry and manipulating texture for more expressiveness.
      Our expression regularization is important to produce realistic geometric expressions, and directional loss for balanced manipulation of both texture and geometry.       }
    \label{fig:expression_ours}
\end{figure*}

Furthermore, we show results for expression manipulation in Fig.~\ref{fig:expression_ours}.
\OURS{} faithfully deforms face geometry and texture to match a variety of text prompts, where expression regularization maintains plausible geometry and directional loss enables balanced adaptation of geometry and texture.
We refer to the supplemental for more visuals.

\begin{figure*}
    \begin{center}
    \includegraphics[width=0.9\linewidth]{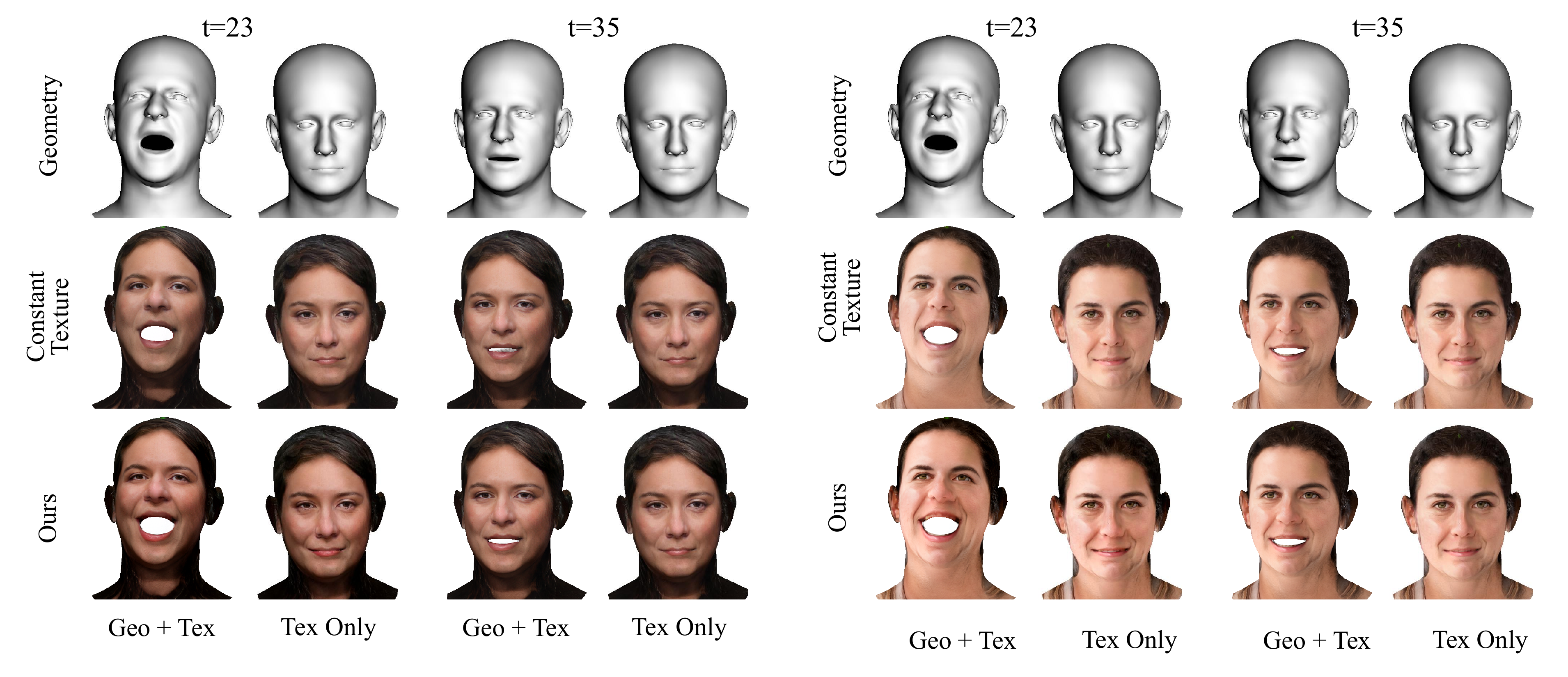}
    \end{center}
    \vspace{-0.6cm}
      \caption{
      Given a 3D face motion sequence (top row), we compare our  dynamic texturing approach (bottom row) against static-only texturing (middle row). Geo + Tex shows textures overlaid on the animated mesh, and Tex Only shows texture in the neutral pose. Our proposed dynamic texture manipulation technique generates more compelling animation, particularly in articulated expressions (e.g., t=23). {Here, we show results for the text prompt "Laughing". We further refer interested readers to our supplemental video.}   
      }
    \label{fig:clip_video_manipualtion}
\end{figure*}

\paragraph{\textbf{Texture Manipulation for Video Sequences}}
Finally, we show results for texture manipulation for given animation sequences in Fig.~\ref{fig:clip_video_manipualtion}. \OURS{} can produce more expressive animation compared to a constant texture that looks monotonic. We show results for only 3 frames; however, we refer readers to the supplemental video for more detailed results.


\paragraph{\textbf{Limitations}}
Although \OURS{} can generate high-quality textures and expressions, it still has various limitations. 
For instance, our method does not capture accessories like jewelry, headwear, or eyewear, due to our use of the FLAME~\cite{flame_siggraphAsia2017} model, which does not represent accessories or complex hair. We believe that this could be further improved by augmenting parametric 3D models with artist-designed assets for 3D hair, headwear, or eyewear.

%% file: pages/05_conclusion.tex
\section{Conclusion}
In this paper, we have introduced \OURS{}, a novel approach to enable text-guided editing of textured 3D morphable face models.
We jointly synthesize high-quality textures and adapt geometry based on the expressions of the morphable model, in a self-supervised fashion.
This enables compelling 3D face generation across a variety of textures, expressions, and styles, based on user-friendly text prompts. 
We further demonstrate the ability of \OURS{} to synthesize of animation sequences, driven by a guiding video sequence.
We believe this is an important first step towards enabling controllable, realistic texture and expression modeling for 3D face models, dovetailing with conventional graphics pipelines, which will enable many new possibilities for content creation and digital avatars.

\section*{Acknowledgments}
This work was supported by the ERC Starting Grant Scan2CAD (804724), the Bavarian State Ministry of Science and the Arts and coordinated by the Bavarian Research Institute for Digital Transformation (bidt), the German Research Foundation (DFG) Grant ``Making Machine Learning on Static and Dynamic 3D Data Practical,'' the German Research Foundation (DFG) Research Unit ``Learning and Simulation in Visual Computing,'' and Sony Semiconductor Solutions Corporation. We would like to thank Yawar Siddiqui for help with implementation of differential rendering codebase, Artem Sevastopolsky and Guy Gafni for helpful discussions, and Alexey Bokhovkin, Chandan Yeshwanth, and Yueh-Cheng Liu for help during internal review.

%% file: pages/06_appendix.tex
\section*{Appendix}

We provide additional ablation studies and results in Section ~\ref{sec:experiments}, network architecture and training details in Section ~\ref{sec:architecture}, and further discussion of baseline method experimental setup in Section ~\ref{sec:baselines}.


\section{Additional Results }\label{sec:experiments}
We provide additional results for texture and expression manipulation, as well as an additional comparison to texturing baselines. 
For results related to video animation, we refer readers to the supplementary video.

\begin{figure*}[h!]
    \begin{center}
    \includegraphics[width=1.0\linewidth]{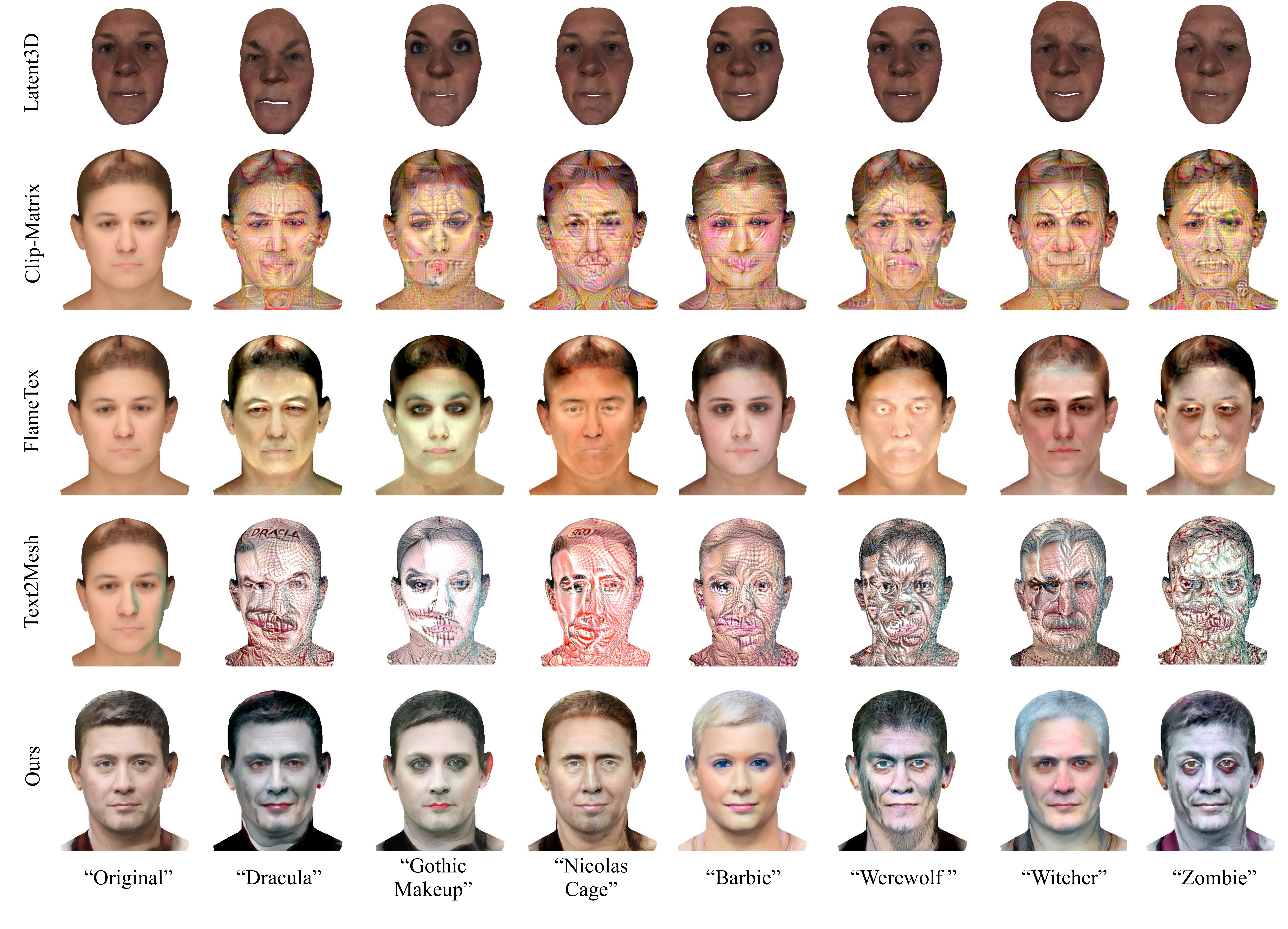}
    \end{center}
    \vspace{-0.5cm}
      \caption{Additional texture manipulation results in comparison with baselines, from the text prompts shown. Our method outperforms baseline approaches in texture quality.
      }
    \label{fig:baseline_comp_supp}
\end{figure*}

\paragraph{\textbf{Additional Baseline Comparisons: }}
We compare our method with texturing baselines for additional text prompts, and show results in Figure~\ref{fig:baseline_comp_supp}. Our method outperforms these baselines and achieves high quality manipulations.

\begin{figure*}[h!]
    \begin{center}
    \includegraphics[width=1.0\linewidth]{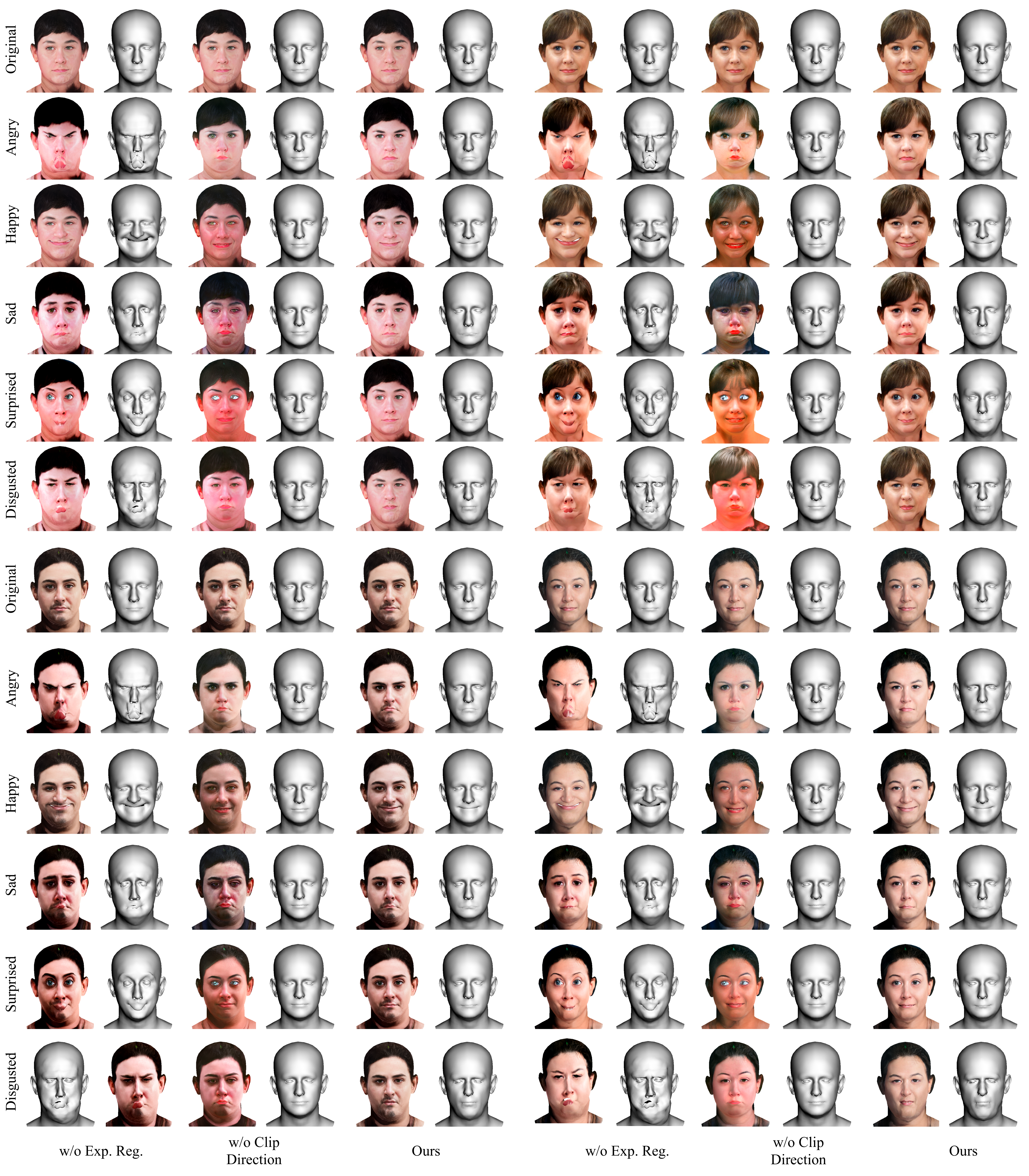}
    \end{center}
    \vspace{-0.5cm}
      \caption{Additional expression manipulation results: our proposed method achieves realistic expression manipulation.
      Both our expression regularization and directional clip loss contribute notably to realistic output quality.
      }
    \label{fig:expression_ablation_supp}
\end{figure*}

\paragraph{\textbf{Expression Manipulation: }}
We show additional results for expression manipulation, and analyze the effect of our expression regularization and directional clip loss in Figure~\ref{fig:expression_ablation_supp}. Our proposed technique outperforms others and achieves realistic texture manipulation.

\begin{figure*}
    \centering
    \begin{subfigure}{0.75\linewidth}
      \centering
      \includegraphics[width=\linewidth]{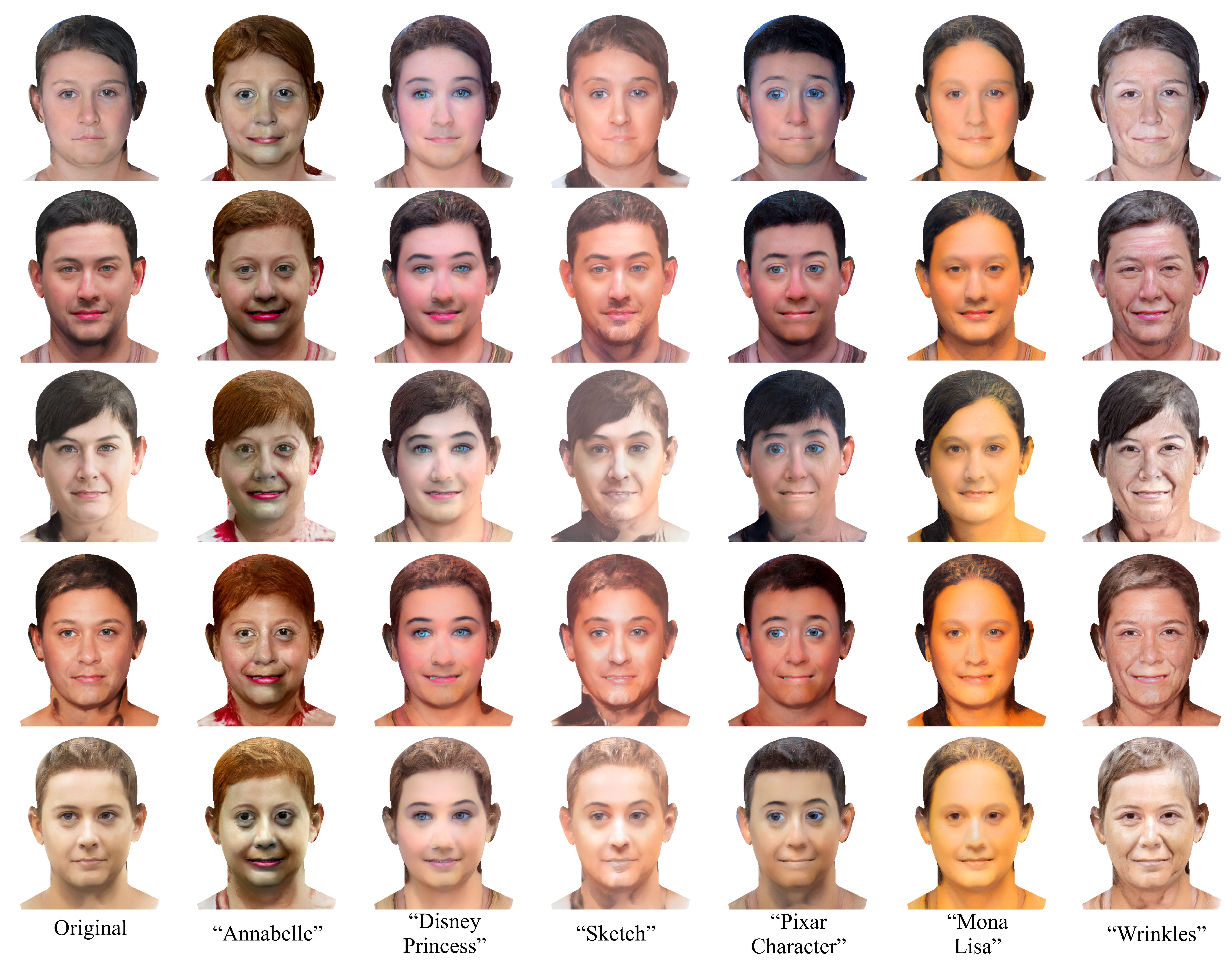}
      \label{fig:sub1}
    \end{subfigure}%
    \hfill
    \begin{subfigure}{0.75\linewidth}
      \centering
      \includegraphics[width=\linewidth]{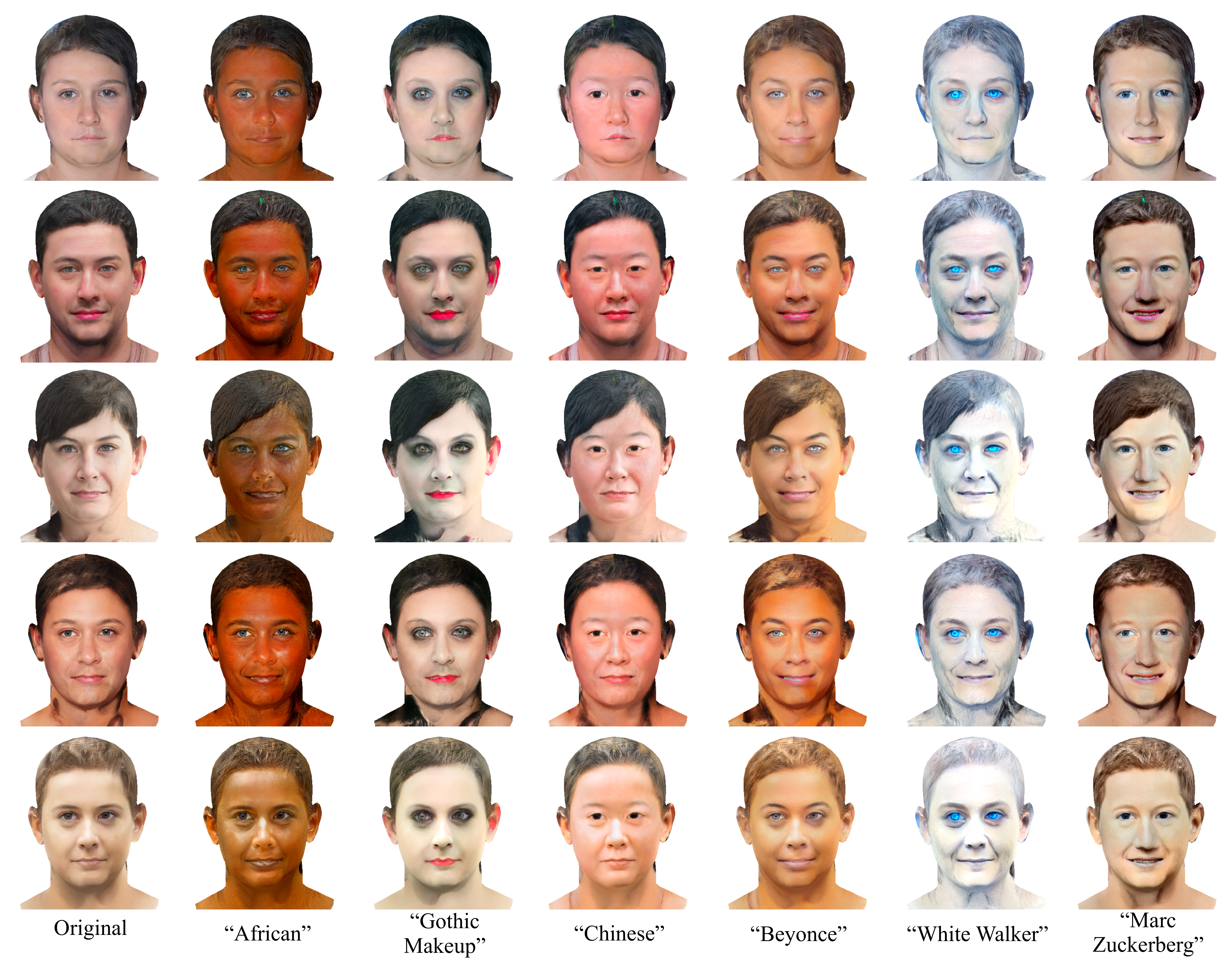}
      \label{fig:sub2}
    \end{subfigure}
    \vspace{-0.8cm}
    \caption{Additional texturing results: Our proposed method generates a diverse range of textures.}
    \label{fig:additional_texturing}
\end{figure*}

\paragraph{\textbf{Texture Manipulation: }}
We show additional texture manipulation results with a large variety of text prompts in Figure~\ref{fig:additional_texturing}.
As can be seen, our method is able to generate a wide variety of textures, even capable of adapting identity when implied by the text prompt.

\begin{figure*}[h!]
    \begin{center}
    \includegraphics[width=1.0\linewidth]{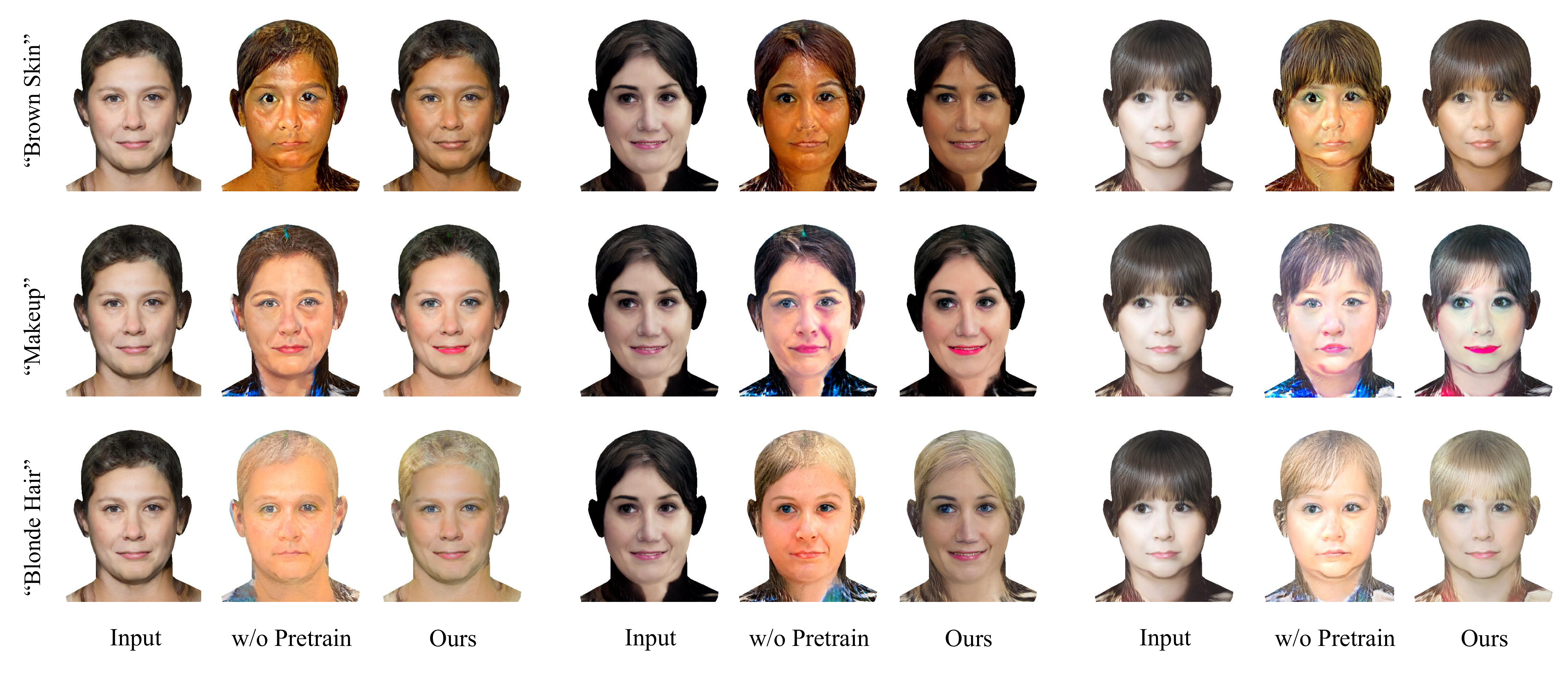}
    \end{center}
    \vspace{-0.5cm}
      \caption{We evaluate the effect of pre-training texture and expression mappers. `w/o Pretrain' refers to the case when mappers are not trained to predict zero offsets before performing text manipulation. Pre-training helps to produce realistic texture changes.
      }
    \label{fig:texture_ablation}
\end{figure*}

\paragraph{\textbf{Effect of Pre-training: }}
Here, we analyze the effect of pre-training the texture and expression mappers. We first pre-train the texture mapper $\mathcal{T}$ and expression mapper $\mathcal{E}$ to predict zero offsets using $\ell_2$ regularization before training them with our CLIP loss. We show the effect of pre-training mapper networks to predict zero offsets in Figure~\ref{fig:texture_ablation}. Without pre-training, the textures begin with unrealistic values and converge to low quality styles with visible artifacts.

\begin{figure*}[h!]
    \begin{center}
    \includegraphics[width=0.75\linewidth]{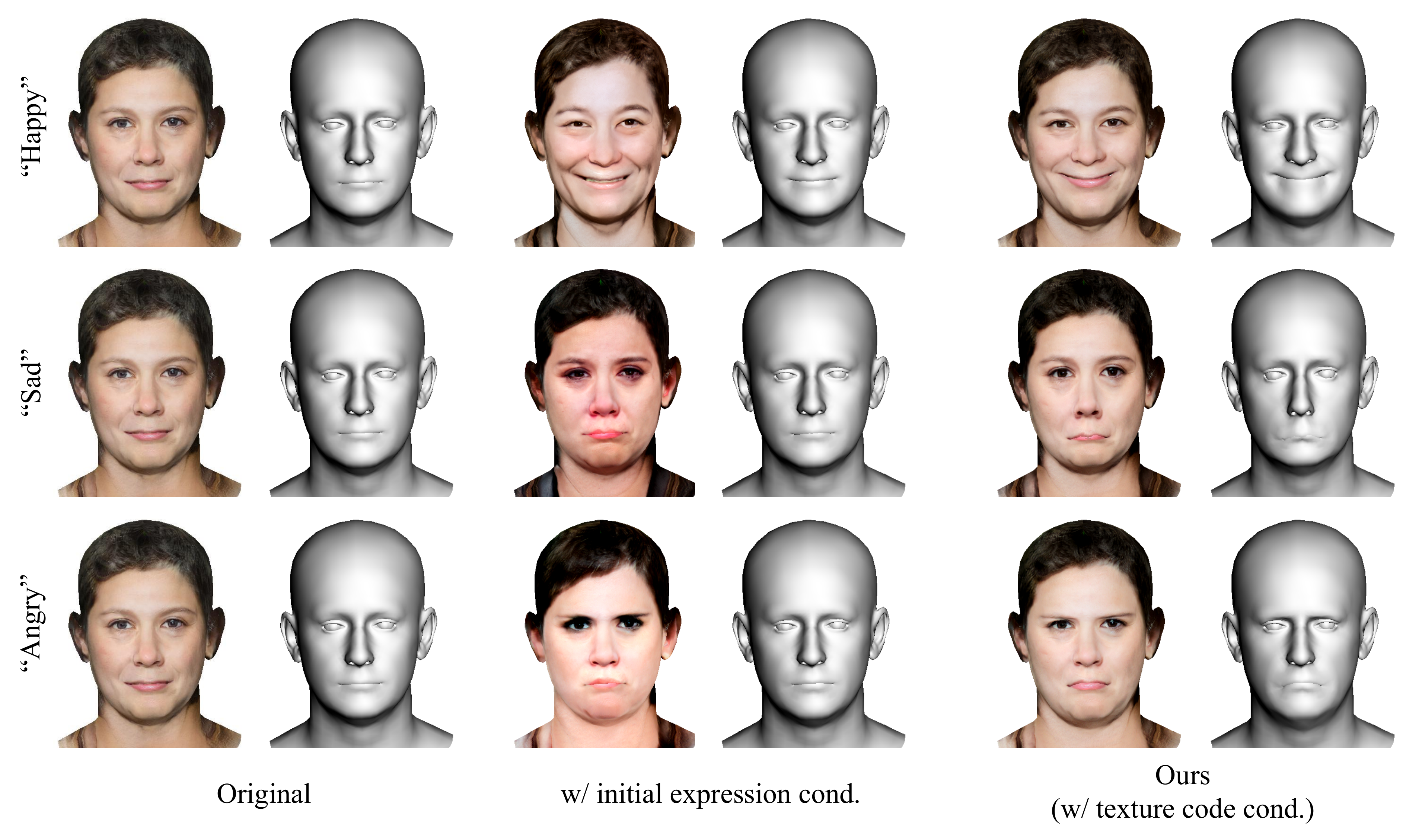}
    \end{center}
    \vspace{-0.5cm}
      \todo{\caption{We evaluate the effect of different conditioning inputs on the expression mapper $\mathcal{E}$ during expression manipulation. `w/ initial expression cond' refers to the case when expression mapper is conditioned on initial expression code $\boldsymbol{\psi}_{\textrm{init}}$, `w/ texture code cond' refers to the case when $\mathcal{E}$ is conditioned on mean texture code $\mathbf{w}_{\textrm{mean}}$. Conditioning on $\mathbf{w}_{\textrm{mean}}$ provides the meaningful signal to modify texture and expression cohesively. 
      }
    \label{fig:input_conditioning_ablation}}
\end{figure*}

{ \paragraph{\textbf{Effect of input conditioning for Expression Mapper: }}
Finally, we analyze the effect of different input conditions on the generated expressions during expression manipulation. Conditioning the expression mapper $\mathcal{E}$ on initial expression code $\boldsymbol{\psi}_{\textrm{init}}$ generates uncanny textures, does not noticeably alter the geometry and attempts to encode all information into the texture. Our proposed approach conditions the $\mathcal{E}$ on mean texture code $\mathbf{w}_{\textrm{mean}}$ providing a meaningful signal to modify texture and expression cohesively. Results are shown in Figure~\ref{fig:input_conditioning_ablation}}.

\begin{figure*}[h!]
    \begin{center}
    \includegraphics[width=0.8\linewidth]{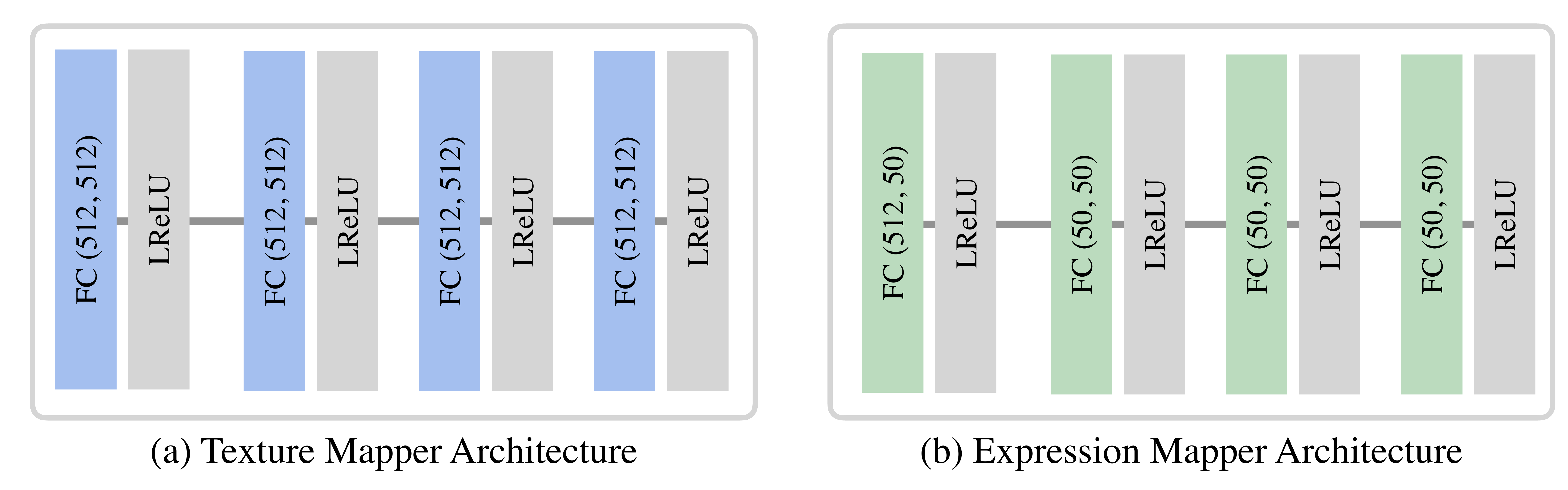}
    \end{center}
    \vspace{-0.5cm}
      \caption{Architecture Overview: Network architecture for Texture Mapper (left) and Expression mapper (right). FC($x,y$) refers to a fully-connected layer, where $x$ and $y$ denote the input and output dimensions, respectively. LReLU refers to LeakyReLU activations.
      }
    \label{fig:model_arch}
\end{figure*}


\section{Architecture \& Training Details}\label{sec:architecture}
ClipFace is implemented in the Pytorch Lightning framework~\cite{paszke2017automatic, falcon2019pytorch}.
For differentiable rendering, we use the NvDiffrast~\cite{Laine2020diffrast} library.

\paragraph{\textbf{Texture Generation:}}
For texture generation, we use the StyleGAN2 architecture with adaptive discriminator augmentation~\cite{Karras2020ada}.
For all of our experiments, we operate at a texture and image resolution of $512 \times 512$.
We apply augmentation to both the full-image discriminator as well as the patch discriminator, which operates on patches of size $64 \times 64$.
For augmentations, we apply geometric transformations such as image flipping, rotation and scaling, as well as color transformations such as changing image brightness, contrast, hue, saturation, etc.
For training, we used the Adam~\cite{adam} optimizer with a learning rate of 0.002, batch size 8 per GPU, gradient penalty 10, and path length regularization 2.
We perform multi-GPU training on 3 RTX A6000 GPUs and train for 300,000 iterations.
%

\paragraph{\textbf{Texture and Expression Manipulation:}}
For the text-guided manipulation experiments, we use a 4-layer MLP architecture with LReLU activations.
For the texture mapper, we use 18 identical MLPs to predict texture offsets for different levels of the latent code $\mathbf{w}_{\textrm{init}} = \{\mathbf{w}_{\textrm{init}}^1, \mathbf{w}_{\textrm{init}}^2, ... \mathbf{w}_{\textrm{init}}^{18} \} \in \mathbb{R}^{512 \times 18}$.
\begin{equation}
    \mathbf{w}_{\textrm{delta}}= 
    \begin{cases}
        \mathcal{T}^{1}(\mathbf{w}_\textrm{{init}}^{1})\\
        \mathcal{T}^{2}(\mathbf{w}_{\textrm{init}}^{2})\\
        \vdots\\
        \mathcal{T}^{18}(\mathbf{w}_{\textrm{init}}^{18}).\\
    \end{cases}
\end{equation}
Each texture MLP $\mathcal{T}^i$ takes as input 512-dimensional latent code $\mathbf{w}_\textrm{{init}}^{i}$ and outputs the 512-dimensional offset $\mathbf{w}_\textrm{{delta}}^{i}$.
The expression mapper $\mathcal{E}$ takes the mean latent code as input $\mathbf{w}_{\textrm{mean}} \in \mathbb{R}^{512}$, and predicts the expression offset $\boldsymbol{\psi}_{\textrm{delta}} \in \mathbb{R}^{50}$ as output.
The network architecture for both these mappers is shown in Figure~\ref{fig:model_arch}.
%
%
%
For language supervision, we use the CLIP model~\cite{clip_radford21a}. For our experiments, we use the pre-trained `ViT-B/32' variant for computing the CLIP loss. We use a learning rate of 0.005 for the expression mapper and 0.0001 for the texture mapper.

\paragraph{\textbf{Texturing for Animation Sequences:}}
For the task of texturing for animations, we learn only the texture mapper.
We use the same architecture as shown in Figure~\ref{fig:model_arch}(a).
Since a given animation sequence consists of multiple frames, we share the texture mapper across different timestamps.
For all text manipulation experiments, we train for 20,000 iterations.
%


\section{Baseline Implementations}\label{sec:baselines}

\paragraph{\textbf{Latent3d~\cite{latent_3d}: }}
This method builds upon the TB-GAN~\cite{gecer2020tbgan}, a generative model $\mathcal{G}$ that takes one-hot encoded facial expression vector $\vec{\mathbf{e}}$ and a random noise vector $\vec{\mathbf{z}} \in \mathbb{R}^{d} $ as input and generates shape, shape-normal and texture images.
Given a pretrained generator $\mathcal{G}$, the method optimizes offset $\Delta \mathbf{c}$ for the intermediate layer $\mathbf{c}$ which is $4 \times 4$ dense layer of TB-GAN.
The offset $\Delta \mathbf{c}$ gives the direction in which the target attributes specified by text prompt $\mathbf{t}$ are enhanced, while other attributes stay unchanged.
The authors use a Clip-loss $\mathcal{L}_{\textrm{CLIP}}$, supplemented with an identity loss $\mathcal{L}_{\textrm{ID}}$ and L2 regularization $\mathcal{L}_{\textrm{L2}}$ to perform meaningful manipulation of meshes:
\begin{equation}
    \argmin_{\Delta \textbf{c} \in \mathcal{C}} \mathcal{L}_{\textrm{CLIP}} + \lambda_{\textrm{ID}} \mathcal{L}_{\textrm{ID}} + \lambda_{\textrm{L2}} \mathcal{L}_{\textrm{L2}},
\end{equation}
where $\lambda_{\textrm{ID}}$ and $\lambda_{\textrm{L2}}$ are hyperparameters for the $\mathcal{L}_{\textrm{ID}}$ and $\mathcal{L}_{\textrm{L2}}$ respectively.
The identity loss $\mathcal{L}_{\textrm{ID}}$ minimizes the distance between the identity of original renders and manipulated renders:
\begin{equation}
    \mathcal{L}_{\textrm{ID}} = 1 - \langle R(\mathcal{G}(\mathbf{c})), R(\mathcal{G}(\mathbf{c} + \Delta  \mathbf{c})) \rangle ,
\end{equation}
where $R$ is the ArcFace~\cite{arcface}, a facial recognition network and $\langle .,.\rangle$ computes the cosine similarity between the identities of the initial rendering and manipulated rendering. The L2 loss is used by authors to prevent artifact generation and can be written as:
\begin{equation}
    \mathcal{L}_{\textrm{L2}} = \| \mathbf{c} - (\mathbf{c} + \Delta \mathbf{c})   \|_2
\end{equation}
For the Clip loss, the authors use a list of text templates like `a photo of a ...'; `a face of a ...', etc prefixed to the target style:
\begin{equation}
    \mathcal{L}_{\textrm{CLIP}} = \frac{\Sigma_{j=1}^K \Sigma_{i=1}^N D_{\textrm{CLIP}} (\mathcal{I}_i, t_j ) }{K \cdot N},
\end{equation}
where $\mathcal{I}_i$ is the rendered image from a list of N rendered images, $t_j$ is the target text $t$ embedded in a text template from a list of K templates.
$D_{\textrm{CLIP}}$ minimizes the cosine distance between CLIP embeddings of the rendered image $\mathcal{I}_i$ and the set of text prompts $t_j$.

\paragraph{\textbf{ClipMatrix~\cite{clip_matrix}:}}
Given a 3D mesh and initial texture map $T_{init}$,  ClipMatrix optimizes the texture image offset $T_{delta}$ to match the image rendering $I$ to the text prompt $t$ from random camera view $c$:
\begin{equation}
    \mathcal{L}(T_{delta}) = \sum_{t}  \underset{c \sim \pi_c}{\mathbb{E}} \mathcal{L}_{\mathrm{CLIP}} (I, t).
\end{equation}
By sampling from random camera angles $c \sim \pi_c$ during optimization, the method ensures that output mesh shows the desired properties from different viewing angles. The image rendering can be obtained as:
\begin{equation}
    I = \mathcal{R}( \mathcal{M}, T_{tgt}, c),
\end{equation}
where $\mathcal{M}$ refers to the 3D mesh, $T_{tgt} = T_{init} + T_{delta}$ denotes the final UV texture map and $c$ denotes the camera view. The clip loss $\mathcal{L}_{\mathrm{CLIP}} (I, t)$ minimizes the negative cosine similarity in CLIP embedding space between image $I$ and the fixed text prompt $t$. 
\begin{equation}
    \mathcal{L}_{\mathrm{CLIP}} (I, t) = - cos (\phi_i(I), \phi_t(t))
\end{equation}
where $\phi_i$ and $\phi_t$ refer to the clip image and text encoder respectively. The texture offset $T_{delta}$ is initialized with zero and during optimization clipping is applied to final texture image $T_{tgt} \in [-1, 1]$ to ensure that it stays in valid image range.

\paragraph{\textbf{FlameTex~\cite{flame_tex}: }}
FlameTex is the PCA-based texturing model designed specifically for FLAME face model~\cite{flame_siggraphAsia2017}. 
The texture space for FlameTex is built using randomly selected 1500 images from the FFHQ dataset~\cite{Karras2018stylegan} and the base texture from the Basel Face Model~\cite{Paysan2009A3F}. 
Given a mean texture $T_{mean} \in \mathbb{R}^{512 \times 512 \times 3}$ and texture basis $T_{basis} \in \mathbb{R}^{50 \times 512 \times 512 \times 3}$ from the FlameTex texture model, we optimize for the texture basis coefficients $\boldsymbol{\omega} \in \mathbb{R}^{50}$ to match the target text prompt $t$ to generate the desired texture map $T_{tgt}$:
\begin{equation}
    \boldsymbol{\omega}^{*} = \argmin_{\boldsymbol{\omega}} \mathcal{L}_{\textrm{CLIP}}(I, t) + \lambda_{\mathrm{L2}} \mathcal{L}_{\mathrm{L2}} (\boldsymbol{\omega}) ,
\end{equation}
where $\mathcal{L}_{\textrm{CLIP}}$ refers to the clip loss between text prompt $t$ and the rendered image $I$ and $\mathcal{L}_{\mathrm{L2}}$ refers to the L2 regularization for the texture coefficients $\boldsymbol{\omega}$ with $\lambda_{\mathrm{L2}}$ controlling the strength of the regularization. The desired texture map is given by:
\begin{equation}
    T_{tgt} = T_{mean} + {\omega} * T_{basis} .
\end{equation}
The image rendering can be obtained as:
\begin{equation}
    I = \mathcal{R}(\mathcal{M}, T_{tgt}, c) ,
\end{equation}
where $\mathcal{M}$ refers to the Flame 3D mesh, $T_{tgt}$ denotes the final UV texture map and $c$ denotes the camera view.
The L2 regularization is applied to prevent the model from producing unrealistic texture and is given by:
\begin{equation}
    \mathcal{L}_{\textrm{L2}} = \| \boldsymbol{\omega} \|_2 .
\end{equation}
We initialize $\boldsymbol{\omega}$ with zero and start from base texture $T_{mean}$ during optimization.

\paragraph{\textbf{Text2Mesh~\cite{Michel_2022_CVPR}: }}
Given a 3D mesh, the method uses coordinate-based MLPs to predict per-vertex color and displacement conforming to the target text prompt $t$ used for stylizing the mesh.
In our experiments, we first remesh the Flame 3D mesh to increase vertices from 5K to 60K as the method works reasonably well for meshes with a higher vertex count.
We used the default hyperparameters used by authors for stylizing human body meshes, as the authors did not perform experiments on human face meshes.
For every vertex point $p$, first the positional encoding $\mathcal{\gamma}$ is applied to obtain high frequency features, before passing them to the MLP:
\begin{equation}
    \gamma(p) = [\textrm{cos} (2\pi \mathbf{B}p), \textrm{sin} (2\pi \mathbf{B}p)]^{T} ,
\end{equation}
where $\mathbf{B} \in \mathbb{R}^{n \times 3}$ is the random Gaussian matrix.
We first pretrain the MLP $f_{\boldsymbol{\theta}}$ to predict a base texture $T_{init}$ to begin learning from a reasonable starting texture.
Since we do not wish to change the geometry, we do not perform vertex displacements in our experiments.
Our sanity experiments for vertex displacements produced unrealistic geometries.
We train with loss function and augmentations proposed in the main paper.
The loss function can be written as:
\begin{equation}
    \boldsymbol{\theta}^{*} = \argmin_{\boldsymbol{\theta}} \mathcal{L}_{\textrm{CLIP}} (I, t) ,
\end{equation}
where $I$ refers to the image rendered from different viewpoints and $t$ refers to the target text prompt.
The image rendering $I$ can be obtained as:
\begin{equation}
    I = \mathcal{R}(\mathcal{M}, [f_{\boldsymbol{\theta}}(\gamma(p_i))]_{i=1}^N, c) ,
\end{equation}
where $\mathcal{R}$ refers to the differentiable rendering, c refers to the random camera view and $p_i \in \mathbb{R}^3$ refers to the vertex coordinate of the mesh and $N$ refers to the total number of mesh vertices.

\clearpage

%% file: main.bbl

\begin{thebibliography}{58}


\ifx \showCODEN    \undefined \def \showCODEN     #1{\unskip}     \fi
\ifx \showDOI      \undefined \def \showDOI       #1{#1}\fi
\ifx \showISBNx    \undefined \def \showISBNx     #1{\unskip}     \fi
\ifx \showISBNxiii \undefined \def \showISBNxiii  #1{\unskip}     \fi
\ifx \showISSN     \undefined \def \showISSN      #1{\unskip}     \fi
\ifx \showLCCN     \undefined \def \showLCCN      #1{\unskip}     \fi
\ifx \shownote     \undefined \def \shownote      #1{#1}          \fi
\ifx \showarticletitle \undefined \def \showarticletitle #1{#1}   \fi
\ifx \showURL      \undefined \def \showURL       {\relax}        \fi
\providecommand\bibfield[2]{#2}
\providecommand\bibinfo[2]{#2}
\providecommand\natexlab[1]{#1}
\providecommand\showeprint[2][]{arXiv:#2}

\bibitem[Abdal et~al\mbox{.}(2021a)]%
        {clip2stylegan}
\bibfield{author}{\bibinfo{person}{Rameen Abdal}, \bibinfo{person}{Peihao Zhu},
  \bibinfo{person}{John Femiani}, \bibinfo{person}{Niloy~J. Mitra}, {and}
  \bibinfo{person}{Peter Wonka}.} \bibinfo{year}{2021}\natexlab{a}.
\newblock \showarticletitle{CLIP2StyleGAN: Unsupervised Extraction of StyleGAN
  Edit Directions}.
\newblock \bibinfo{journal}{\emph{CoRR}}  \bibinfo{volume}{abs/2112.05219}
  (\bibinfo{year}{2021}).
\newblock
\showeprint[arXiv]{2112.05219}
\urldef\tempurl%
\url{https://arxiv.org/abs/2112.05219}
\showURL{%
\tempurl}


\bibitem[Abdal et~al\mbox{.}(2021b)]%
        {styleflow}
\bibfield{author}{\bibinfo{person}{Rameen Abdal}, \bibinfo{person}{Peihao Zhu},
  \bibinfo{person}{Niloy~J. Mitra}, {and} \bibinfo{person}{Peter Wonka}.}
  \bibinfo{year}{2021}\natexlab{b}.
\newblock \showarticletitle{StyleFlow: Attribute-Conditioned Exploration of
  StyleGAN-Generated Images Using Conditional Continuous Normalizing Flows}.
\newblock \bibinfo{journal}{\emph{ACM Trans. Graph.}} (\bibinfo{date}{May}
  \bibinfo{year}{2021}).
\newblock
\showISSN{0730-0301}
\urldef\tempurl%
\url{https://doi.org/10.1145/3447648}
\showDOI{\tempurl}


\bibitem[Avrahami et~al\mbox{.}(2022)]%
        {avrahami2022blended}
\bibfield{author}{\bibinfo{person}{Omri Avrahami}, \bibinfo{person}{Dani
  Lischinski}, {and} \bibinfo{person}{Ohad Fried}.}
  \bibinfo{year}{2022}\natexlab{}.
\newblock \showarticletitle{Blended diffusion for text-driven editing of
  natural images}. In \bibinfo{booktitle}{\emph{Proceedings of the IEEE/CVF
  Conference on Computer Vision and Pattern Recognition}}.
  \bibinfo{pages}{18208--18218}.
\newblock


\bibitem[Bau et~al\mbox{.}(2021)]%
        {bau2021paintbyword}
\bibfield{author}{\bibinfo{person}{David Bau}, \bibinfo{person}{Alex Andonian},
  \bibinfo{person}{Audrey Cui}, \bibinfo{person}{YeonHwan Park},
  \bibinfo{person}{Ali Jahanian}, \bibinfo{person}{Aude Oliva}, {and}
  \bibinfo{person}{Antonio Torralba}.} \bibinfo{year}{2021}\natexlab{}.
\newblock \bibinfo{title}{Paint by Word}.
\newblock
\newblock
\showeprint{arXiv:2103.10951}


\bibitem[Blanz and Vetter(1999)]%
        {3dmm}
\bibfield{author}{\bibinfo{person}{Volker Blanz} {and} \bibinfo{person}{Thomas
  Vetter}.} \bibinfo{year}{1999}\natexlab{}.
\newblock \showarticletitle{A Morphable Model for the Synthesis of 3D Faces}.
  In \bibinfo{booktitle}{\emph{Proceedings of the 26th Annual Conference on
  Computer Graphics and Interactive Techniques}}
  \emph{(\bibinfo{series}{SIGGRAPH '99})}. \bibinfo{publisher}{ACM
  Press/Addison-Wesley Publishing Co.}, \bibinfo{address}{USA},
  \bibinfo{pages}{187–194}.
\newblock
\showISBNx{0201485605}
\urldef\tempurl%
\url{https://doi.org/10.1145/311535.311556}
\showDOI{\tempurl}


\bibitem[Canfes et~al\mbox{.}(2022)]%
        {latent_3d}
\bibfield{author}{\bibinfo{person}{Zehranaz Canfes}, \bibinfo{person}{M.~Furkan
  Atasoy}, \bibinfo{person}{Alara Dirik}, {and} \bibinfo{person}{Pinar
  Yanardag}.} \bibinfo{year}{2022}\natexlab{}.
\newblock \bibinfo{title}{Text and Image Guided 3D Avatar Generation and
  Manipulation}.
\newblock
\newblock
\urldef\tempurl%
\url{https://doi.org/10.48550/ARXIV.2202.06079}
\showDOI{\tempurl}


\bibitem[Chan et~al\mbox{.}(2021)]%
        {Chan2021}
\bibfield{author}{\bibinfo{person}{Eric~R. Chan}, \bibinfo{person}{Connor~Z.
  Lin}, \bibinfo{person}{Matthew~A. Chan}, \bibinfo{person}{Koki Nagano},
  \bibinfo{person}{Boxiao Pan}, \bibinfo{person}{Shalini~De Mello},
  \bibinfo{person}{Orazio Gallo}, \bibinfo{person}{Leonidas Guibas},
  \bibinfo{person}{Jonathan Tremblay}, \bibinfo{person}{Sameh Khamis},
  \bibinfo{person}{Tero Karras}, {and} \bibinfo{person}{Gordon Wetzstein}.}
  \bibinfo{year}{2021}\natexlab{}.
\newblock \showarticletitle{Efficient Geometry-aware {3D} Generative
  Adversarial Networks}. In \bibinfo{booktitle}{\emph{arXiv}}.
\newblock


\bibitem[Crowson(2021)]%
        {vqgan_clip}
\bibfield{author}{\bibinfo{person}{Katherine Crowson}.}
  \bibinfo{year}{2021}\natexlab{}.
\newblock \bibinfo{title}{VQGAN-CLIP}.
\newblock
\newblock
\urldef\tempurl%
\url{https://github.com/nerdyrodent/VQGAN-CLIP}
\showURL{%
\tempurl}


\bibitem[Dayma et~al\mbox{.}(2021)]%
        {Dayma_dalle_Mini_2021}
\bibfield{author}{\bibinfo{person}{Boris Dayma}, \bibinfo{person}{Suraj Patil},
  \bibinfo{person}{Pedro Cuenca}, \bibinfo{person}{Khalid Saifullah},
  \bibinfo{person}{Tanishq Abraham}, \bibinfo{person}{Phuc Le~Khac},
  \bibinfo{person}{Luke Melas}, {and} \bibinfo{person}{Ritobrata Ghosh}.}
  \bibinfo{year}{2021}\natexlab{}.
\newblock \bibinfo{title}{DALL·E Mini}.
\newblock
\newblock
\urldef\tempurl%
\url{https://doi.org/10.5281/zenodo.5146400}
\showDOI{\tempurl}


\bibitem[Deng et~al\mbox{.}(2019)]%
        {arcface}
\bibfield{author}{\bibinfo{person}{Jiankang Deng}, \bibinfo{person}{Jia Guo},
  \bibinfo{person}{Niannan Xue}, {and} \bibinfo{person}{Stefanos Zafeiriou}.}
  \bibinfo{year}{2019}\natexlab{}.
\newblock \showarticletitle{ArcFace: Additive Angular Margin Loss for Deep Face
  Recognition}. In \bibinfo{booktitle}{\emph{2019 IEEE/CVF Conference on
  Computer Vision and Pattern Recognition (CVPR)}}.
  \bibinfo{pages}{4685--4694}.
\newblock
\urldef\tempurl%
\url{https://doi.org/10.1109/CVPR.2019.00482}
\showDOI{\tempurl}


\bibitem[Deng et~al\mbox{.}(2020)]%
        {deng2020disentangled}
\bibfield{author}{\bibinfo{person}{Yu Deng}, \bibinfo{person}{Jiaolong Yang},
  \bibinfo{person}{Dong Chen}, \bibinfo{person}{Fang Wen}, {and}
  \bibinfo{person}{Xin Tong}.} \bibinfo{year}{2020}\natexlab{}.
\newblock \showarticletitle{Disentangled and Controllable Face Image Generation
  via 3D Imitative-Contrastive Learning}. In \bibinfo{booktitle}{\emph{IEEE
  Computer Vision and Pattern Recognition}}.
\newblock


\bibitem[Falcon et~al\mbox{.}(2019)]%
        {falcon2019pytorch}
\bibfield{author}{\bibinfo{person}{William Falcon} {et~al\mbox{.}}}
  \bibinfo{year}{2019}\natexlab{}.
\newblock \showarticletitle{Pytorch lightning}.
\newblock \bibinfo{journal}{\emph{GitHub. Note: https://github.
  com/PyTorchLightning/pytorch-lightning}} \bibinfo{volume}{3},
  \bibinfo{number}{6} (\bibinfo{year}{2019}).
\newblock


\bibitem[Feng(2019)]%
        {flame_tex}
\bibfield{author}{\bibinfo{person}{Haven Feng}.}
  \bibinfo{year}{2019}\natexlab{}.
\newblock \bibinfo{title}{Photometric FLAME Fitting}.
\newblock
  \bibinfo{howpublished}{\url{https://github.com/HavenFeng/photometric_optimization}}.
\newblock


\bibitem[Feng et~al\mbox{.}(2021)]%
        {deca}
\bibfield{author}{\bibinfo{person}{Yao Feng}, \bibinfo{person}{Haiwen Feng},
  \bibinfo{person}{Michael~J. Black}, {and} \bibinfo{person}{Timo Bolkart}.}
  \bibinfo{year}{2021}\natexlab{}.
\newblock \showarticletitle{Learning an Animatable Detailed {3D} Face Model
  from In-the-Wild Images}.
\newblock \bibinfo{journal}{\emph{ACM Transactions on Graphics (ToG), Proc.
  SIGGRAPH}} \bibinfo{volume}{40}, \bibinfo{number}{4} (\bibinfo{date}{Aug.}
  \bibinfo{year}{2021}), \bibinfo{pages}{88:1--88:13}.
\newblock


\bibitem[Gal et~al\mbox{.}(2021)]%
        {gal2021stylegannada}
\bibfield{author}{\bibinfo{person}{Rinon Gal}, \bibinfo{person}{Or Patashnik},
  \bibinfo{person}{Haggai Maron}, \bibinfo{person}{Gal Chechik}, {and}
  \bibinfo{person}{Daniel Cohen-Or}.} \bibinfo{year}{2021}\natexlab{}.
\newblock \bibinfo{title}{StyleGAN-NADA: CLIP-Guided Domain Adaptation of Image
  Generators}.
\newblock
\newblock
\showeprint[arxiv]{2108.00946}~[cs.CV]


\bibitem[Gecer et~al\mbox{.}(2021a)]%
        {Gecer_2021_CVPR}
\bibfield{author}{\bibinfo{person}{Baris Gecer}, \bibinfo{person}{Jiankang
  Deng}, {and} \bibinfo{person}{Stefanos Zafeiriou}.}
  \bibinfo{year}{2021}\natexlab{a}.
\newblock \showarticletitle{OSTeC: One-Shot Texture Completion}. In
  \bibinfo{booktitle}{\emph{Proceedings of the IEEE/CVF Conference on Computer
  Vision and Pattern Recognition (CVPR)}}. \bibinfo{pages}{7628--7638}.
\newblock


\bibitem[{Gecer} et~al\mbox{.}(2020)]%
        {gecer2020tbgan}
\bibfield{author}{\bibinfo{person}{Baris {Gecer}}, \bibinfo{person}{Alexander
  {Lattas}}, \bibinfo{person}{Stylianos {Ploumpis}}, \bibinfo{person}{Jiankang
  {Deng}}, \bibinfo{person}{Athanasios {Papaioannou}},
  \bibinfo{person}{Stylianos {Moschoglou}}, {and} \bibinfo{person}{Stefanos
  {Zafeiriou}}.} \bibinfo{year}{2020}\natexlab{}.
\newblock \showarticletitle{Synthesizing Coupled 3D Face Modalities by
  Trunk-Branch Generative Adversarial Networks}. In
  \bibinfo{booktitle}{\emph{Proceedings of the European conference on computer
  vision (ECCV)}}. Springer.
\newblock


\bibitem[Gecer et~al\mbox{.}(2019)]%
        {Gecer_2019_CVPR}
\bibfield{author}{\bibinfo{person}{Baris Gecer}, \bibinfo{person}{Stylianos
  Ploumpis}, \bibinfo{person}{Irene Kotsia}, {and} \bibinfo{person}{Stefanos
  Zafeiriou}.} \bibinfo{year}{2019}\natexlab{}.
\newblock \showarticletitle{GANFIT: Generative Adversarial Network Fitting for
  High Fidelity 3D Face Reconstruction}. In
  \bibinfo{booktitle}{\emph{Proceedings of the IEEE/CVF Conference on Computer
  Vision and Pattern Recognition (CVPR)}}.
\newblock


\bibitem[Gecer et~al\mbox{.}(2021b)]%
        {gecer2021fast}
\bibfield{author}{\bibinfo{person}{Baris Gecer}, \bibinfo{person}{Stylianos
  Ploumpis}, \bibinfo{person}{Irene Kotsia}, {and} \bibinfo{person}{Stefanos~P
  Zafeiriou}.} \bibinfo{year}{2021}\natexlab{b}.
\newblock \showarticletitle{Fast-GANFIT: Generative Adversarial Network for
  High Fidelity 3D Face Reconstruction}.
\newblock \bibinfo{journal}{\emph{IEEE Transactions on Pattern Analysis and
  Machine Intelligence}} (\bibinfo{year}{2021}).
\newblock


\bibitem[Gerig et~al\mbox{.}(2017)]%
        {bfm_17}
\bibfield{author}{\bibinfo{person}{Thomas Gerig}, \bibinfo{person}{Andreas
  Morel-Forster}, \bibinfo{person}{Clemens Blumer}, \bibinfo{person}{Bernhard
  Egger}, \bibinfo{person}{Marcel Lüthi}, \bibinfo{person}{Sandro Schönborn},
  {and} \bibinfo{person}{Thomas Vetter}.} \bibinfo{year}{2017}\natexlab{}.
\newblock \bibinfo{title}{Morphable Face Models - An Open Framework}.
\newblock
\newblock
\urldef\tempurl%
\url{https://doi.org/10.48550/ARXIV.1709.08398}
\showDOI{\tempurl}


\bibitem[Ghosh et~al\mbox{.}(2020)]%
        {GIF2020}
\bibfield{author}{\bibinfo{person}{Partha Ghosh}, \bibinfo{person}{Pravir~Singh
  Gupta}, \bibinfo{person}{Roy Uziel}, \bibinfo{person}{Anurag Ranjan},
  \bibinfo{person}{Michael~J. Black}, {and} \bibinfo{person}{Timo Bolkart}.}
  \bibinfo{year}{2020}\natexlab{}.
\newblock \showarticletitle{{GIF}: Generative Interpretable Faces}. In
  \bibinfo{booktitle}{\emph{International Conference on 3D Vision (3DV)}}.
  \bibinfo{pages}{868--878}.
\newblock
\urldef\tempurl%
\url{http://gif.is.tue.mpg.de/}
\showURL{%
\tempurl}


\bibitem[Hong et~al\mbox{.}(2022)]%
        {hong2022avatarclip}
\bibfield{author}{\bibinfo{person}{Fangzhou Hong}, \bibinfo{person}{Mingyuan
  Zhang}, \bibinfo{person}{Liang Pan}, \bibinfo{person}{Zhongang Cai},
  \bibinfo{person}{Lei Yang}, {and} \bibinfo{person}{Ziwei Liu}.}
  \bibinfo{year}{2022}\natexlab{}.
\newblock \showarticletitle{AvatarCLIP: Zero-Shot Text-Driven Generation and
  Animation of 3D Avatars}.
\newblock \bibinfo{journal}{\emph{ACM Transactions on Graphics (TOG)}}
  \bibinfo{volume}{41}, \bibinfo{number}{4} (\bibinfo{year}{2022}),
  \bibinfo{pages}{1--19}.
\newblock


\bibitem[Jetchev(2021)]%
        {clip_matrix}
\bibfield{author}{\bibinfo{person}{Nikolay Jetchev}.}
  \bibinfo{year}{2021}\natexlab{}.
\newblock \bibinfo{title}{ClipMatrix: Text-controlled Creation of 3D Textured
  Meshes}.
\newblock
\newblock
\urldef\tempurl%
\url{https://doi.org/10.48550/ARXIV.2109.12922}
\showDOI{\tempurl}


\bibitem[Karras et~al\mbox{.}(2018)]%
        {karras2018progressive}
\bibfield{author}{\bibinfo{person}{Tero Karras}, \bibinfo{person}{Timo Aila},
  \bibinfo{person}{Samuli Laine}, {and} \bibinfo{person}{Jaakko Lehtinen}.}
  \bibinfo{year}{2018}\natexlab{}.
\newblock \showarticletitle{Progressive Growing of {GAN}s for Improved Quality,
  Stability, and Variation}. In \bibinfo{booktitle}{\emph{International
  Conference on Learning Representations}}.
\newblock
\urldef\tempurl%
\url{https://openreview.net/forum?id=Hk99zCeAb}
\showURL{%
\tempurl}


\bibitem[Karras et~al\mbox{.}(2020a)]%
        {Karras2020ada}
\bibfield{author}{\bibinfo{person}{Tero Karras}, \bibinfo{person}{Miika
  Aittala}, \bibinfo{person}{Janne Hellsten}, \bibinfo{person}{Samuli Laine},
  \bibinfo{person}{Jaakko Lehtinen}, {and} \bibinfo{person}{Timo Aila}.}
  \bibinfo{year}{2020}\natexlab{a}.
\newblock \showarticletitle{Training Generative Adversarial Networks with
  Limited Data}. In \bibinfo{booktitle}{\emph{Proc. NeurIPS}}.
\newblock


\bibitem[Karras et~al\mbox{.}(2019)]%
        {Karras2018stylegan}
\bibfield{author}{\bibinfo{person}{Tero Karras}, \bibinfo{person}{Samuli
  Laine}, {and} \bibinfo{person}{Timo Aila}.} \bibinfo{year}{2019}\natexlab{}.
\newblock \showarticletitle{A Style-Based Generator Architecture for Generative
  Adversarial Networks}. In \bibinfo{booktitle}{\emph{2019 IEEE/CVF Conference
  on Computer Vision and Pattern Recognition (CVPR)}}.
  \bibinfo{pages}{4396--4405}.
\newblock
\urldef\tempurl%
\url{https://doi.org/10.1109/CVPR.2019.00453}
\showDOI{\tempurl}


\bibitem[Karras et~al\mbox{.}(2020b)]%
        {Karras2019stylegan2}
\bibfield{author}{\bibinfo{person}{Tero Karras}, \bibinfo{person}{Samuli
  Laine}, \bibinfo{person}{Miika Aittala}, \bibinfo{person}{Janne Hellsten},
  \bibinfo{person}{Jaakko Lehtinen}, {and} \bibinfo{person}{Timo Aila}.}
  \bibinfo{year}{2020}\natexlab{b}.
\newblock \showarticletitle{Analyzing and Improving the Image Quality of
  {StyleGAN}}. In \bibinfo{booktitle}{\emph{Proc. CVPR}}.
\newblock


\bibitem[Khalid et~al\mbox{.}(2022)]%
        {khalid2022clipmesh}
\bibfield{author}{\bibinfo{person}{Nasir~Mohammad Khalid},
  \bibinfo{person}{Tianhao Xie}, \bibinfo{person}{Eugene Belilovsky}, {and}
  \bibinfo{person}{Popa Tiberiu}.} \bibinfo{year}{2022}\natexlab{}.
\newblock \showarticletitle{CLIP-Mesh: Generating textured meshes from text
  using pretrained image-text models}.
\newblock  (\bibinfo{date}{December} \bibinfo{year}{2022}).
\newblock


\bibitem[Kingma and Ba(2014)]%
        {adam}
\bibfield{author}{\bibinfo{person}{Diederik~P. Kingma} {and}
  \bibinfo{person}{Jimmy Ba}.} \bibinfo{year}{2014}\natexlab{}.
\newblock \bibinfo{title}{Adam: A Method for Stochastic Optimization}.
\newblock
\newblock
\urldef\tempurl%
\url{https://doi.org/10.48550/ARXIV.1412.6980}
\showDOI{\tempurl}


\bibitem[Kocasari et~al\mbox{.}(2022)]%
        {Kocasari_2022_WACV}
\bibfield{author}{\bibinfo{person}{Umut Kocasari}, \bibinfo{person}{Alara
  Dirik}, \bibinfo{person}{Mert Tiftikci}, {and} \bibinfo{person}{Pinar
  Yanardag}.} \bibinfo{year}{2022}\natexlab{}.
\newblock \showarticletitle{StyleMC: Multi-Channel Based Fast Text-Guided Image
  Generation and Manipulation}. In \bibinfo{booktitle}{\emph{Proceedings of the
  IEEE/CVF Winter Conference on Applications of Computer Vision (WACV)}}.
  \bibinfo{pages}{895--904}.
\newblock


\bibitem[Kowalski et~al\mbox{.}(2020)]%
        {KowalskiECCV2020}
\bibfield{author}{\bibinfo{person}{Marek Kowalski}, \bibinfo{person}{Stephan~J.
  Garbin}, \bibinfo{person}{Virginia Estellers}, \bibinfo{person}{Tadas
  Baltrušaitis}, \bibinfo{person}{Matthew Johnson}, {and}
  \bibinfo{person}{Jamie Shotton}.} \bibinfo{year}{2020}\natexlab{}.
\newblock \showarticletitle{CONFIG: Controllable Neural Face Image Generation}.
  In \bibinfo{booktitle}{\emph{European Conference on Computer Vision (ECCV)}}.
\newblock


\bibitem[Laine et~al\mbox{.}(2020)]%
        {Laine2020diffrast}
\bibfield{author}{\bibinfo{person}{Samuli Laine}, \bibinfo{person}{Janne
  Hellsten}, \bibinfo{person}{Tero Karras}, \bibinfo{person}{Yeongho Seol},
  \bibinfo{person}{Jaakko Lehtinen}, {and} \bibinfo{person}{Timo Aila}.}
  \bibinfo{year}{2020}\natexlab{}.
\newblock \showarticletitle{Modular Primitives for High-Performance
  Differentiable Rendering}.
\newblock \bibinfo{journal}{\emph{ACM Transactions on Graphics}}
  \bibinfo{volume}{39}, \bibinfo{number}{6} (\bibinfo{year}{2020}).
\newblock


\bibitem[Lattas et~al\mbox{.}(2020)]%
        {lattas2020avatarme}
\bibfield{author}{\bibinfo{person}{Alexandros Lattas},
  \bibinfo{person}{Stylianos Moschoglou}, \bibinfo{person}{Baris Gecer},
  \bibinfo{person}{Stylianos Ploumpis}, \bibinfo{person}{Vasileios
  Triantafyllou}, \bibinfo{person}{Abhijeet Ghosh}, {and}
  \bibinfo{person}{Stefanos Zafeiriou}.} \bibinfo{year}{2020}\natexlab{}.
\newblock \showarticletitle{AvatarMe: Realistically Renderable 3D Facial
  Reconstruction "In-the-Wild"}. In \bibinfo{booktitle}{\emph{Proceedings of
  the IEEE/CVF Conference on Computer Vision and Pattern Recognition (CVPR)}}.
\newblock


\bibitem[Lattas et~al\mbox{.}(2021)]%
        {lattas2021avatarme++}
\bibfield{author}{\bibinfo{person}{Alexandros Lattas},
  \bibinfo{person}{Stylianos Moschoglou}, \bibinfo{person}{Stylianos Ploumpis},
  \bibinfo{person}{Baris Gecer}, \bibinfo{person}{Abhijeet Ghosh}, {and}
  \bibinfo{person}{Stefanos~P Zafeiriou}.} \bibinfo{year}{2021}\natexlab{}.
\newblock \showarticletitle{AvatarMe++: Facial Shape and BRDF Inference with
  Photorealistic Rendering-Aware GANs}.
\newblock \bibinfo{journal}{\emph{IEEE Transactions on Pattern Analysis and
  Machine Intelligence}} (\bibinfo{year}{2021}).
\newblock


\bibitem[Lee et~al\mbox{.}(2020)]%
        {Lee2020StyleUVDA}
\bibfield{author}{\bibinfo{person}{Myunggi Lee}, \bibinfo{person}{Wonwoong
  Cho}, \bibinfo{person}{Moonheum Kim}, \bibinfo{person}{David~I. Inouye},
  {and} \bibinfo{person}{Nojun Kwak}.} \bibinfo{year}{2020}\natexlab{}.
\newblock \showarticletitle{StyleUV: Diverse and High-fidelity UV Map
  Generative Model}.
\newblock \bibinfo{journal}{\emph{ArXiv}}  \bibinfo{volume}{abs/2011.12893}
  (\bibinfo{year}{2020}).
\newblock


\bibitem[Li et~al\mbox{.}(2020)]%
        {Li_2020_CVPR}
\bibfield{author}{\bibinfo{person}{Ruilong Li}, \bibinfo{person}{Karl Bladin},
  \bibinfo{person}{Yajie Zhao}, \bibinfo{person}{Chinmay Chinara},
  \bibinfo{person}{Owen Ingraham}, \bibinfo{person}{Pengda Xiang},
  \bibinfo{person}{Xinglei Ren}, \bibinfo{person}{Pratusha Prasad},
  \bibinfo{person}{Bipin Kishore}, \bibinfo{person}{Jun Xing}, {and}
  \bibinfo{person}{Hao Li}.} \bibinfo{year}{2020}\natexlab{}.
\newblock \showarticletitle{Learning Formation of Physically-Based Face
  Attributes}. In \bibinfo{booktitle}{\emph{IEEE/CVF Conference on Computer
  Vision and Pattern Recognition (CVPR)}}.
\newblock


\bibitem[Li et~al\mbox{.}(2017)]%
        {flame_siggraphAsia2017}
\bibfield{author}{\bibinfo{person}{Tianye Li}, \bibinfo{person}{Timo Bolkart},
  \bibinfo{person}{Michael.~J. Black}, \bibinfo{person}{Hao Li}, {and}
  \bibinfo{person}{Javier Romero}.} \bibinfo{year}{2017}\natexlab{}.
\newblock \showarticletitle{Learning a model of facial shape and expression
  from {4D} scans}.
\newblock \bibinfo{journal}{\emph{ACM Transactions on Graphics, (Proc. SIGGRAPH
  Asia)}} \bibinfo{volume}{36}, \bibinfo{number}{6} (\bibinfo{year}{2017}),
  \bibinfo{pages}{194:1--194:17}.
\newblock


\bibitem[Liu et~al\mbox{.}(2022)]%
        {Liu20223DFMGT}
\bibfield{author}{\bibinfo{person}{Yuchen Liu}, \bibinfo{person}{Zhixin Shu},
  \bibinfo{person}{Yijun Li}, \bibinfo{person}{Zhe Lin},
  \bibinfo{person}{Richard Zhang}, {and} \bibinfo{person}{S.~Y. Kung}.}
  \bibinfo{year}{2022}\natexlab{}.
\newblock \showarticletitle{3D-FM GAN: Towards 3D-Controllable Face
  Manipulation}.
\newblock \bibinfo{journal}{\emph{ArXiv}}  \bibinfo{volume}{abs/2208.11257}
  (\bibinfo{year}{2022}).
\newblock


\bibitem[Loper et~al\mbox{.}(2015)]%
        {SMPL_2015}
\bibfield{author}{\bibinfo{person}{Matthew Loper}, \bibinfo{person}{Naureen
  Mahmood}, \bibinfo{person}{Javier Romero}, \bibinfo{person}{Gerard
  Pons-Moll}, {and} \bibinfo{person}{Michael~J. Black}.}
  \bibinfo{year}{2015}\natexlab{}.
\newblock \showarticletitle{{SMPL}: A Skinned Multi-Person Linear Model}.
\newblock \bibinfo{journal}{\emph{ACM Trans. Graphics (Proc. SIGGRAPH Asia)}}
  \bibinfo{volume}{34}, \bibinfo{number}{6} (\bibinfo{date}{Oct.}
  \bibinfo{year}{2015}), \bibinfo{pages}{248:1--248:16}.
\newblock


\bibitem[Luo et~al\mbox{.}(2021)]%
        {Luo_2021_CVPR}
\bibfield{author}{\bibinfo{person}{Huiwen Luo}, \bibinfo{person}{Koki Nagano},
  \bibinfo{person}{Han-Wei Kung}, \bibinfo{person}{Qingguo Xu},
  \bibinfo{person}{Zejian Wang}, \bibinfo{person}{Lingyu Wei},
  \bibinfo{person}{Liwen Hu}, {and} \bibinfo{person}{Hao Li}.}
  \bibinfo{year}{2021}\natexlab{}.
\newblock \showarticletitle{Normalized Avatar Synthesis Using StyleGAN and
  Perceptual Refinement}. In \bibinfo{booktitle}{\emph{Proceedings of the
  IEEE/CVF Conference on Computer Vision and Pattern Recognition (CVPR)}}.
  \bibinfo{pages}{11662--11672}.
\newblock


\bibitem[Marriott et~al\mbox{.}(2021)]%
        {Marriott2021A3G}
\bibfield{author}{\bibinfo{person}{Richard~T. Marriott}, \bibinfo{person}{Sami
  Romdhani}, {and} \bibinfo{person}{Liming Chen}.}
  \bibinfo{year}{2021}\natexlab{}.
\newblock \showarticletitle{A 3D GAN for Improved Large-pose Facial
  Recognition}.
\newblock \bibinfo{journal}{\emph{2021 IEEE/CVF Conference on Computer Vision
  and Pattern Recognition (CVPR)}} (\bibinfo{year}{2021}),
  \bibinfo{pages}{13440--13450}.
\newblock


\bibitem[Michel et~al\mbox{.}(2022)]%
        {Michel_2022_CVPR}
\bibfield{author}{\bibinfo{person}{Oscar Michel}, \bibinfo{person}{Roi Bar-On},
  \bibinfo{person}{Richard Liu}, \bibinfo{person}{Sagie Benaim}, {and}
  \bibinfo{person}{Rana Hanocka}.} \bibinfo{year}{2022}\natexlab{}.
\newblock \showarticletitle{Text2Mesh: Text-Driven Neural Stylization for
  Meshes}. In \bibinfo{booktitle}{\emph{Proceedings of the IEEE/CVF Conference
  on Computer Vision and Pattern Recognition (CVPR)}}.
  \bibinfo{pages}{13492--13502}.
\newblock


\bibitem[Paszke et~al\mbox{.}(2017)]%
        {paszke2017automatic}
\bibfield{author}{\bibinfo{person}{Adam Paszke}, \bibinfo{person}{Sam Gross},
  \bibinfo{person}{Soumith Chintala}, \bibinfo{person}{Gregory Chanan},
  \bibinfo{person}{Edward Yang}, \bibinfo{person}{Zachary DeVito},
  \bibinfo{person}{Zeming Lin}, \bibinfo{person}{Alban Desmaison},
  \bibinfo{person}{Luca Antiga}, {and} \bibinfo{person}{Adam Lerer}.}
  \bibinfo{year}{2017}\natexlab{}.
\newblock \showarticletitle{Automatic differentiation in PyTorch}.
\newblock  (\bibinfo{year}{2017}).
\newblock


\bibitem[Patashnik et~al\mbox{.}(2021)]%
        {Patashnik_2021_ICCV}
\bibfield{author}{\bibinfo{person}{Or Patashnik}, \bibinfo{person}{Zongze Wu},
  \bibinfo{person}{Eli Shechtman}, \bibinfo{person}{Daniel Cohen-Or}, {and}
  \bibinfo{person}{Dani Lischinski}.} \bibinfo{year}{2021}\natexlab{}.
\newblock \showarticletitle{StyleCLIP: Text-Driven Manipulation of StyleGAN
  Imagery}. In \bibinfo{booktitle}{\emph{Proceedings of the IEEE/CVF
  International Conference on Computer Vision (ICCV)}}.
  \bibinfo{pages}{2085--2094}.
\newblock


\bibitem[Paysan et~al\mbox{.}(2009)]%
        {Paysan2009A3F}
\bibfield{author}{\bibinfo{person}{Pascal Paysan}, \bibinfo{person}{Reinhard
  Knothe}, \bibinfo{person}{Brian Amberg}, \bibinfo{person}{Sami Romdhani},
  {and} \bibinfo{person}{Thomas Vetter}.} \bibinfo{year}{2009}\natexlab{}.
\newblock \showarticletitle{A 3D Face Model for Pose and Illumination Invariant
  Face Recognition}.
\newblock \bibinfo{journal}{\emph{2009 Sixth IEEE International Conference on
  Advanced Video and Signal Based Surveillance}} (\bibinfo{year}{2009}),
  \bibinfo{pages}{296--301}.
\newblock


\bibitem[Petrovich et~al\mbox{.}(2022)]%
        {petrovich22temos}
\bibfield{author}{\bibinfo{person}{Mathis Petrovich},
  \bibinfo{person}{Michael~J. Black}, {and} \bibinfo{person}{G{\"u}l Varol}.}
  \bibinfo{year}{2022}\natexlab{}.
\newblock \showarticletitle{{TEMOS}: Generating diverse human motions from
  textual descriptions}. In \bibinfo{booktitle}{\emph{European Conference on
  Computer Vision ({ECCV})}}.
\newblock


\bibitem[Radford et~al\mbox{.}(2021)]%
        {clip_radford21a}
\bibfield{author}{\bibinfo{person}{Alec Radford}, \bibinfo{person}{Jong~Wook
  Kim}, \bibinfo{person}{Chris Hallacy}, \bibinfo{person}{Aditya Ramesh},
  \bibinfo{person}{Gabriel Goh}, \bibinfo{person}{Sandhini Agarwal},
  \bibinfo{person}{Girish Sastry}, \bibinfo{person}{Amanda Askell},
  \bibinfo{person}{Pamela Mishkin}, \bibinfo{person}{Jack Clark},
  \bibinfo{person}{Gretchen Krueger}, {and} \bibinfo{person}{Ilya Sutskever}.}
  \bibinfo{year}{2021}\natexlab{}.
\newblock \showarticletitle{Learning Transferable Visual Models From Natural
  Language Supervision}. In \bibinfo{booktitle}{\emph{Proceedings of the 38th
  International Conference on Machine Learning}}
  \emph{(\bibinfo{series}{Proceedings of Machine Learning Research},
  Vol.~\bibinfo{volume}{139})}, \bibfield{editor}{\bibinfo{person}{Marina
  Meila} {and} \bibinfo{person}{Tong Zhang}} (Eds.). \bibinfo{publisher}{PMLR},
  \bibinfo{pages}{8748--8763}.
\newblock
\urldef\tempurl%
\url{https://proceedings.mlr.press/v139/radford21a.html}
\showURL{%
\tempurl}


\bibitem[Ramesh et~al\mbox{.}(2022)]%
        {dalle2}
\bibfield{author}{\bibinfo{person}{Aditya Ramesh}, \bibinfo{person}{Prafulla
  Dhariwal}, \bibinfo{person}{Alex Nichol}, \bibinfo{person}{Casey Chu}, {and}
  \bibinfo{person}{Mark Chen}.} \bibinfo{year}{2022}\natexlab{}.
\newblock \bibinfo{title}{Hierarchical Text-Conditional Image Generation with
  CLIP Latents}.
\newblock
\newblock
\urldef\tempurl%
\url{https://doi.org/10.48550/ARXIV.2204.06125}
\showDOI{\tempurl}


\bibitem[Ramesh et~al\mbox{.}(2021)]%
        {dalle}
\bibfield{author}{\bibinfo{person}{Aditya Ramesh}, \bibinfo{person}{Mikhail
  Pavlov}, \bibinfo{person}{Gabriel Goh}, \bibinfo{person}{Scott Gray},
  \bibinfo{person}{Chelsea Voss}, \bibinfo{person}{Alec Radford},
  \bibinfo{person}{Mark Chen}, {and} \bibinfo{person}{Ilya Sutskever}.}
  \bibinfo{year}{2021}\natexlab{}.
\newblock \bibinfo{title}{Zero-Shot Text-to-Image Generation}.
\newblock
\newblock
\urldef\tempurl%
\url{https://doi.org/10.48550/ARXIV.2102.12092}
\showDOI{\tempurl}


\bibitem[Ruiz et~al\mbox{.}(2022)]%
        {ruiz2022dreambooth}
\bibfield{author}{\bibinfo{person}{Nataniel Ruiz}, \bibinfo{person}{Yuanzhen
  Li}, \bibinfo{person}{Varun Jampani}, \bibinfo{person}{Yael Pritch},
  \bibinfo{person}{Michael Rubinstein}, {and} \bibinfo{person}{Kfir Aberman}.}
  \bibinfo{year}{2022}\natexlab{}.
\newblock \showarticletitle{DreamBooth: Fine Tuning Text-to-image Diffusion
  Models for Subject-Driven Generation}.
\newblock  (\bibinfo{year}{2022}).
\newblock


\bibitem[Slossberg et~al\mbox{.}(2022)]%
        {slossberg2021unsupervised}
\bibfield{author}{\bibinfo{person}{Ron Slossberg}, \bibinfo{person}{Ibrahim
  Jubran}, {and} \bibinfo{person}{Ron Kimmel}.}
  \bibinfo{year}{2022}\natexlab{}.
\newblock \showarticletitle{Unsupervised High-Fidelity Facial Texture
  Generation and Reconstruction}. In \bibinfo{booktitle}{\emph{Computer Vision
  – ECCV 2022: 17th European Conference, Tel Aviv, Israel, October 23–27,
  2022, Proceedings, Part XIII}} (Tel Aviv, Israel).
  \bibinfo{publisher}{Springer-Verlag}, \bibinfo{address}{Berlin, Heidelberg},
  \bibinfo{pages}{212–229}.
\newblock
\showISBNx{978-3-031-19777-2}
\urldef\tempurl%
\url{https://doi.org/10.1007/978-3-031-19778-9_13}
\showDOI{\tempurl}


\bibitem[Tewari et~al\mbox{.}(2020a)]%
        {tewari2020stylerig}
\bibfield{author}{\bibinfo{person}{Ayush Tewari}, \bibinfo{person}{Mohamed
  Elgharib}, \bibinfo{person}{Gaurav Bharaj}, \bibinfo{person}{Florian
  Bernard}, \bibinfo{person}{Hans-Peter Seidel}, \bibinfo{person}{Patrick
  P{\'e}rez}, \bibinfo{person}{Michael Z{\"o}llhofer}, {and}
  \bibinfo{person}{Christian Theobalt}.} \bibinfo{year}{2020}\natexlab{a}.
\newblock \showarticletitle{StyleRig: Rigging StyleGAN for 3D Control over
  Portrait Images, CVPR 2020}. In \bibinfo{booktitle}{\emph{{IEEE} Conference
  on Computer Vision and Pattern Recognition (CVPR)}}. {IEEE}.
\newblock


\bibitem[Tewari et~al\mbox{.}(2020b)]%
        {tewari2020pie}
\bibfield{author}{\bibinfo{person}{Ayush Tewari}, \bibinfo{person}{Mohamed
  Elgharib}, \bibinfo{person}{Mallikarjun~B R}, \bibinfo{person}{Florian
  Bernard}, \bibinfo{person}{Hans-Peter Seidel}, \bibinfo{person}{Patrick
  P\'{e}rez}, \bibinfo{person}{Michael Zollh\"{o}fer}, {and}
  \bibinfo{person}{Christian Theobalt}.} \bibinfo{year}{2020}\natexlab{b}.
\newblock \showarticletitle{PIE: Portrait Image Embedding for Semantic
  Control}.
\newblock \bibinfo{journal}{\emph{ACM Trans. Graph.}} (\bibinfo{year}{2020}).
\newblock


\bibitem[Wang et~al\mbox{.}(2022a)]%
        {Wang_2022_CVPR_clip_nerf}
\bibfield{author}{\bibinfo{person}{Can Wang}, \bibinfo{person}{Menglei Chai},
  \bibinfo{person}{Mingming He}, \bibinfo{person}{Dongdong Chen}, {and}
  \bibinfo{person}{Jing Liao}.} \bibinfo{year}{2022}\natexlab{a}.
\newblock \showarticletitle{CLIP-NeRF: Text-and-Image Driven Manipulation of
  Neural Radiance Fields}. In \bibinfo{booktitle}{\emph{Proceedings of the
  IEEE/CVF Conference on Computer Vision and Pattern Recognition (CVPR)}}.
  \bibinfo{pages}{3835--3844}.
\newblock


\bibitem[Wang et~al\mbox{.}(2022b)]%
        {Wang_2022_CVPR}
\bibfield{author}{\bibinfo{person}{Lizhen Wang}, \bibinfo{person}{Zhiyuan
  Chen}, \bibinfo{person}{Tao Yu}, \bibinfo{person}{Chenguang Ma},
  \bibinfo{person}{Liang Li}, {and} \bibinfo{person}{Yebin Liu}.}
  \bibinfo{year}{2022}\natexlab{b}.
\newblock \showarticletitle{FaceVerse: A Fine-Grained and Detail-Controllable
  3D Face Morphable Model From a Hybrid Dataset}. In
  \bibinfo{booktitle}{\emph{Proceedings of the IEEE/CVF Conference on Computer
  Vision and Pattern Recognition (CVPR)}}. \bibinfo{pages}{20333--20342}.
\newblock


\bibitem[Wei et~al\mbox{.}(2022)]%
        {Wei_2022_CVPR}
\bibfield{author}{\bibinfo{person}{Tianyi Wei}, \bibinfo{person}{Dongdong
  Chen}, \bibinfo{person}{Wenbo Zhou}, \bibinfo{person}{Jing Liao},
  \bibinfo{person}{Zhentao Tan}, \bibinfo{person}{Lu Yuan},
  \bibinfo{person}{Weiming Zhang}, {and} \bibinfo{person}{Nenghai Yu}.}
  \bibinfo{year}{2022}\natexlab{}.
\newblock \showarticletitle{HairCLIP: Design Your Hair by Text and Reference
  Image}. In \bibinfo{booktitle}{\emph{Proceedings of the IEEE/CVF Conference
  on Computer Vision and Pattern Recognition (CVPR)}}.
  \bibinfo{pages}{18072--18081}.
\newblock


\bibitem[Youwang et~al\mbox{.}(2022)]%
        {youwang2022clipactor}
\bibfield{author}{\bibinfo{person}{Kim Youwang}, \bibinfo{person}{Kim Ji-Yeon},
  {and} \bibinfo{person}{Tae-Hyun Oh}.} \bibinfo{year}{2022}\natexlab{}.
\newblock \showarticletitle{CLIP-Actor: Text-Driven Recommendation and
  Stylization for Animating Human Meshes}. In \bibinfo{booktitle}{\emph{ECCV}}.
\newblock


\bibitem[zllrunning(2018)]%
        {face_parsing_pytorch}
\bibfield{author}{\bibinfo{person}{zllrunning}.}
  \bibinfo{year}{2018}\natexlab{}.
\newblock \bibinfo{title}{face-parsing.PyTorch}.
\newblock
  \bibinfo{howpublished}{\url{https://github.com/zllrunning/face-parsing.PyTorch}}.
\newblock


\end{thebibliography}
